\documentclass[twocolumn]{svjour3}          
\smartqed  
\usepackage{graphicx}
\usepackage{mathptmx}      
\usepackage{natbib}

\usepackage{amsmath}
\usepackage{amssymb}

\usepackage{algorithm}
\usepackage{algorithmic}
\usepackage{enumerate}

\usepackage{subfigure}
\usepackage{multirow}
\usepackage{hhline}
\usepackage{enumitem}

\begin{document}

\title{Scheduled Differentiable Architecture Search for Visual Recognition}


\author{Zhaofan Qiu \and Ting Yao \and Yiheng Zhang \and Yongdong Zhang \and Tao Mei
}


\institute{Zhaofan Qiu \at
			University of Science and Technology of China, Hefei, China. \\
			\email{zhaofanqiu@gmail.com}
			\and
			Ting Yao \at
			JD AI Research, Beijing, China. \\
			\email{tingyao.ustc@gmail.com}
			\and
			Yiheng Zhang \at
			University of Science and Technology of China, Hefei, China. \\
			\email{yihengzhang.chn@gmail.com}
			\and
			Yongdong Zhang \at
			University of Science and Technology of China, Hefei, China. \\
			\email{zhyd73@ustc.edu.cn}
			\and
			Tao Mei \at
			JD AI Research, Beijing, China. \\
			\email{tmei@live.com}
}

\date{Received: date / Accepted: date}

\maketitle

\begin{abstract}
  Convolutional Neural Networks (CNN) have been regarded as a capable class of models for visual recognition problems. Nevertheless, it is not trivial to develop generic and powerful network architectures, which requires significant efforts of human experts. In this paper, we introduce a new idea for automatically exploring architectures on a remould of Differentiable Architecture Search (DAS), which possesses the efficient search via gradient descent. Specifically, we present Scheduled Differentiable Architecture Search (SDAS) for both image and video recognition that nicely integrates the selection of operations during training with a schedule. Technically, an architecture or a cell is represented as a directed graph. Our SDAS gradually fixes the operations on the edges in the graph in a progressive and scheduled manner, as opposed to a one-step decision of operations for all the edges once the training completes in existing DAS, which may make the architecture brittle. Moreover, we enlarge the search space of SDAS particularly for video recognition by devising several unique operations to encode spatio-temporal dynamics and demonstrate the impact in affecting the architecture search of SDAS. Extensive experiments of architecture learning are conducted on CIFAR10, Kinetics10, UCF101 and HMDB51 datasets, and superior results are reported when comparing to DAS method. More remarkably, the search by our SDAS is around 2-fold faster than DAS. When transferring the learnt cells on CIFAR10 and Kinetics10 respectively to large-scale ImageNet and Kinetics400 datasets, the constructed network also outperforms several state-of-the-art hand-crafted structures.
\end{abstract}

\section{Introduction}
The development of deep neural networks have successfully pushed the limits of visual recognition with remarkable improvements in state-of-the-art performance. For example, the top-5 error of an ensemble of residual nets \citep{he2015deep} decreases to 3.57\% on the ImageNet dataset \citep{russakovsky2015imagenet}, and the first rank performance achieves 10.99\% in terms of the average of top-1 and top-5 errors in trimmed video recognition task of ActivityNet Challenge 2018 \citep{ghanem2018activitynet}. The achievements rest on the basis of the impressive design on the newly-minted 2D or 3D Convolutional Neural Networks (CNN). Nevertheless, developing a powerful and generic network structure often requires significant engineering of human experts. To reduce the efforts on human-invented architectures and speed up the procedure of exploring neural networks particularly on the dataset of interest, there have been several techniques being proposed for automating the architecture design.

There are two general directions along the exploitation of automatic network architecture search: discovering evolution over a discrete search space \citep{liu2017progressive, liu2018hierarchical, real2018regularized, zoph2018learning} and relaxing the search space to be continuous \citep{liu2018darts, xie2018snas}. The former often capitalizes on one controller to sample networks with different structures in a discrete search space, learn the generated networks and in turn update the weights of the controller in the next round till convergence. Such methods demand expensive computations since a large number of evaluations are required. Instead of optimizing over a discrete set of structures, the search space could be relaxed to be continuous and the process of architecture search is then performed by efficient gradient descent optimization. Please also note that the controller is not involved in this case, making the search framework more flexible. We follow this elegant recipe and employ the continuous relaxation of architecture search in our work, which is more fit for the heavy computations on visual data.

\begin{figure}[!tb]
   \centering
   \subfigure[DAS.]{
     \label{fig:das-dis:a}
     \includegraphics[width=0.4\textwidth]{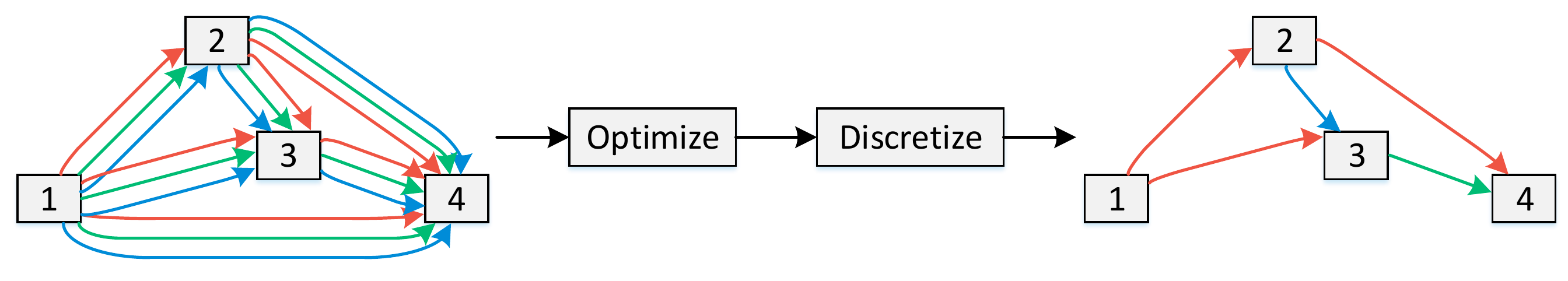}}
   \subfigure[Scheduled DAS (SDAS).]{
     \label{fig:sdas-dis:b}
     \includegraphics[width=0.4\textwidth]{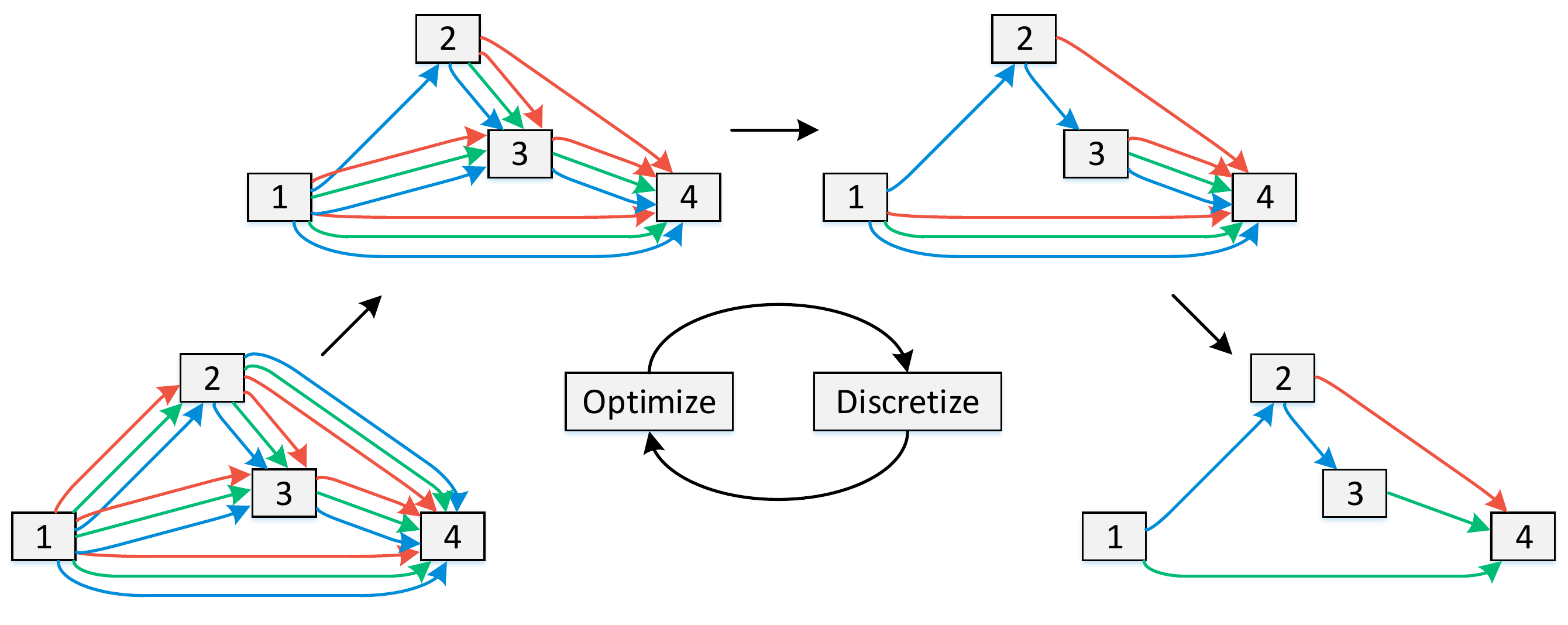}}
   \caption{\small The conceptualization of architecture discretization in (a) DAS and (b) our SDAS.}
    \vspace{-0.15in}
\end{figure}

One typical method to relax the search space is Differentiable Architecture Search (DAS), which represents the architecture (cell) as a directed graph and executes continuous relaxation by weighted mixing all the operations on each edge. The seek for the optimal architecture is then converted to learn the weight of each operation by using gradient descent. Once the learning completes, the architecture is fixed by choosing the only operation with the highest weight as the one on each edge whilst irrationally ignoring the other operations. Figure \ref{fig:das-dis:a} conceptually illustrates the process of architecture discretization in DAS. As such, the learnt architecture may suffer from stability problem. Therefore, we propose to alleviate the problem by progressively inducing the operations on the edges during training with a schedule in order to gradually teach the model to produce a stable architecture, as shown in Figure \ref{fig:sdas-dis:b}. More importantly, considering that video is a temporal sequence which contains both spatial and temporal dimensions, we devise several unique operations, such as convolutions, channel-wise scaling and channel-wise bias, but all in spatio-temporal 3D~mode.

By consolidating the idea of scheduled arrangement of operations in automatic architecture search, we present a new Scheduled Differentiable Architecture Search (SDAS) for visual recognition. Specifically, we depict an architecture or a cell as a directed graph, in which each directed edge is associated with a set of candidate operations. The evolution of the architecture is then equivalent to jointly optimizing the weights or variables of operations on each edge and network parameters through gradient descent. In the training, we employ a schedule, whereby we relax the choice of a particular operation on each edge as a softmax over all the operations at the beginning and gradually discretize the operation on each edge in turn as learning proceeds. Once all the cells have been fixed, we re-train the whole network with the searched cells to fully optimize the network parameters for visual recognition.

The main contribution of this work is the proposal of SDAS to automate the network design for visual recognition. The solution also leads to the elegant view of how to devise unique operations particularly for modeling spatio-temporal dynamics in videos, and how to scheme an effective and efficient strategy for architecture search, which are problems not yet fully understood in the literature.

\section{Related Work}
We briefly group the related works into three categories: image recognition, network architecture search and video recognition.

\textbf{Image Recognition} has received intensive research attention in the area of computer vision and the rise of Convolutional Neural Networks has achieved remarkable performances on several benchmarks. A lot of recent efforts have been made to achieve a high-performing neural network architecture by human experts, such as AlexNet \citep{krizhevsky2012imagenet}, Inception \citep{szegedy2015going}, VGG \citep{Simonyan:ICLR15}, BNInception \citep{ioffe2015batch}, ResNet \citep{he2015deep}, ResNeXt \citep{xie2017aggregated}, Xception \citep{Chollet2017CVPR}, MobileNet \citep{howard2017mobilenets}, ShuffleNet \citep{zhang2018shufflenet} and SENet \citep{hu2018squeeze}. More recently, to automatically design network architectures with less manual intervention, researchers have proposed various \textbf{Network Architecture Search} approaches for image recognition. One successful direction is with reinforcement learning \citep{zoph2017neural}, which devises a controller network to generate the network architecture. Zoph \emph{et al.} \citep{zoph2018learning} further improve the search space in \citep{zoph2017neural} and obtain the state-of-the-art performances on the tasks of image classification and natural language processing. Despite the remarkable performance, this computationally expensive approach takes 1800 GPU days. Several approaches for speeding up the process have been proposed, such as imposing a particular structure of the search space \citep{liu2017progressive, liu2018hierarchical}, weights prediction for each individual architecture \citep{baker2018accelerating, brock2017smash} and weight sharing across multiple architectures \citep{cai2018efficient, pham2018efficient}. Different from the approaches which discover network evolution over a discrete search space, differentiable architecture search \citep{liu2018darts} relaxes the search space to be continuous, that optimizes the architecture by gradient descent and achieves competitive performance using fewer computation resources. Similarly, SNAS~\citep{xie2018snas} reformulates architecture search as an optimization problem of a joint distribution of the search space and make a generic differentiable loss for architecture search.

The early deep models on \textbf{Video Recognition} mostly extend 2D CNN for image recognition to video frames. Karparty \emph{et al.} \citep{karpathy2014large} adapted frame-based 2D CNN for a clip input with fixed temporal window size. The two-stream architecture \citep{simonyan2014two} is devised by utilizing two networks separately on visual frames and stacked optical flow images. This two-stream scheme is further enhanced by exploiting convolutional fusion \citep{feichtenhofer2016convolutional}, key-volume mining \citep{zhu2016key}, temporal segment networks \citep{wang2016temporal,wang2018temporal} and temporal linear encoding \citep{diba2017deep}. The aforementioned approaches often treat a video as a sequence of frames or optical flow images for video recognition. Nevertheless, the pixel-level temporal evolution across consecutive frames are seldom explored. To alleviate this issue, 3D CNN in \citep{ji20133d} is devised to directly learn spatio-temporal representation from a short video clip by 3D convolution. Later in \citep{tran2015learning}, Tran \emph{et al.} design a widely adopted 3D CNN, namely C3D, consisting of 3D convolutions and 3D poolings optimized on large-scale Sports1M \citep{karpathy2014large} dataset. More advanced techniques are studied and proposed recently for 3D CNN, including inflating 2D convolutional kernels \citep{carreira2017quo} and decomposing 3D convolutional kernels \citep{qiu2017learning,tran2018closer}. In this work, we apply neural architecture search for automating 3D CNN backbone design specifically for video recognition.

In short, our work in this paper mainly focuses on the gradient-based architecture search for visual recognition tasks. Different form the previous methods of DARTS \citep{liu2018darts} which chooses all the operations at once after optimization, we propose a scheduled scheme that progressively induces the optimal operation on each edge during training. Moreover, we enlarge the novel search space particularly for video recognition by devising several unique spatio-temporal operations.

\section{Scheduled Differentiable Architecture Search (SDAS)}
The idea of Scheduled Differentiable Architecture Search (SDAS) is to improve the automatic search of an architecture or a cell in a scheduled manner. In this way, SDAS gradually approaches the optimal architectures or cells, making the structure of the whole network more stable. The search procedure can be efficiently implemented by using gradient descent. We begin this Section by presenting the continuous relaxation of search space which enables a joint optimization of architecture and network parameters, and followed by the scheduled discretization scheme to progressively determine the structure of the architecture.

\subsection{Continuous Search Space}
Following the standard practice in the works \citep{liu2017progressive, liu2018hierarchical, liu2018darts, real2018regularized, zoph2018learning} of automatic architecture search, we search for the basic cells and manually predetermine the way of connections between these cells. Derived from the idea of DAS \citep{liu2018darts}, we formulate a computation cell as a directed acyclic graph which consists of an ordered sequence of $N$ nodes. Each node $x^{(i)}$ is a feature map and each directed edge $(j, i)$ is associated with an operation $o^{(j, i)}$. We assume the cell to have two input nodes which are from two prior cells, and the other nodes in the cell are all intermediate ones inferred from the predecessor nodes. Formally, for each intermediate node, the feature map $x^{(i)}$ can be calculated as
\begin{equation}
\small
\begin{aligned}
~~~~~~~~~~ ~~~~~~~~~~ ~~~~~~~~~~ ~~~~~ {x^{(i)}=\sum_{j<i}o^{(j, i)}(x^{(j)})}
\end{aligned}~~,
\label{eq:node}
\end{equation}
where $o^{(j, i)}\in \mathcal{O}$ is the assigned operation on edge $(j, i)$ and $\mathcal{O}$ is the set of candidate operations. The output of each cell is achieved by depth-wisely concatenating all the intermediate nodes. As a result, the problem of architecture search is equivalent to learning the operation on each edge in the directed graph over a discrete search space, which demands expensive computations.

To allow efficient search of the architecture, the choice of a particular operation on each edge is relaxed as a softmax over all the operations. As such, the search space changes to be continuous and the architecture can be optimized with respect to the validation performance by gradient descent. Specifically, the mixing operation $\bar o^{(j, i)}$ on edge $(j, i)$ is formulated as
\begin{equation}
\small
\label{eq:out}
\begin{aligned}
~~~~~~~~~~ ~~~~~~~~~~ ~~~~~~~~~~ \bar o^{(j, i)}(x)=\bar\beta^{(j, i)} \sum_{o\in \mathcal{O}}\bar\alpha_{o}^{(j, i)}o(x)
\end{aligned}~~,
\end{equation}
where $\bar\alpha^{(j, i)}=softmax(\alpha^{(j, i)})$ denotes the learnable weight of different operations on identical edge $(j, i)$, and $\bar\beta^{(j, i)}=sigmoid(\beta^{(j, i)})$ is the effect from node $j$ to node $i$. In this case, the continuous variables $\alpha$ and $\beta$ can be utilized to measure the operation selected on each edge and topological structure of the directed acyclic graph, respectively. Specifically, the predecessor node $j$ with top-$k$ strongest effect $\beta^{(j, i)}$ will be connected to intermediate node $i$, and the final choice of $o^{(j, i)}$ is the operation with highest weight, i.e., $o^{(j, i)}={\arg\max}_{o \in \mathcal{O}}~\alpha_{o}^{(j, i)}$.

By assuming the weights of operations/edges $\{\alpha, \beta\}$ and the parameters $w$ in the operations (e.g., convolutional kernels) are independent of each other as in \citep{liu2018darts}, the two are jointly optimized by
\begin{equation}
\small
\label{eq:loss}
\begin{aligned}
~~~~~~~~~~ ~~~~~~~~~~ ~~~~~~~~~~ \mathcal{L}_{train}(w | \alpha, \beta) + \mathcal{L}_{val}(\alpha, \beta | w)
\end{aligned}~~,
\end{equation}
in which $\mathcal{L}_{train}$ measures the loss on training set for learning the parameters $w$ in the operations, and $\mathcal{L}_{val}$ estimates the loss on validation set for updating the weights $\{\alpha, \beta\}$ of operations/edges in the architecture.

\subsection{Scheduled Discretization} \label{sec:sd}
When performing continuous relaxation of search space and optimizing the architecture by gradient descent, a valid question is then how to obtain an ultimate architecture which is discrete. The natural way is to directly discretize the architecture by choosing the only operation with the highest weight as the one on each edge. The rationale behind is based on the assumption that the validation loss of the architecture in relaxed mode could reflect the performance of the architecture in discrete mode. However, in practice, such discretization scheme once ignores all the other operations, making the change dramatically. Therefore, this kind of scheme may result in skewed optimization of the architecture.

\begin{figure}[!tb]
   \centering {\includegraphics[width=0.33\textwidth]{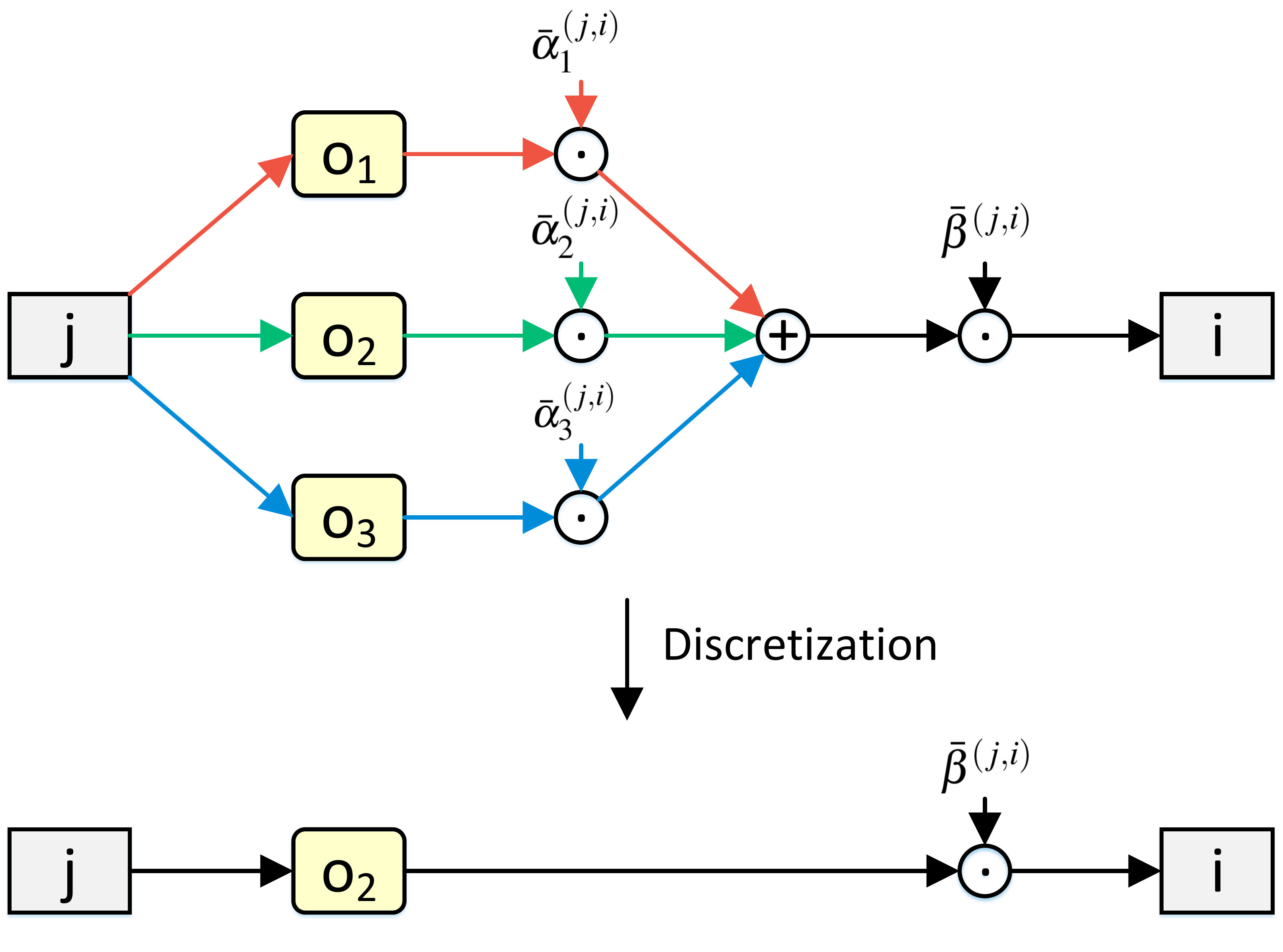}}
   \caption{\small The example of edge discretization. After discretization on edge $(j, i)$, only the operation $o$ with highest weight $\alpha_{o}^{(j, i)}$ is calculated.}
   \label{fig:edge_discrete}
   \vspace{-0.15in}
\end{figure}

\begin{figure}[!tb]
   \centering {\includegraphics[width=0.33\textwidth]{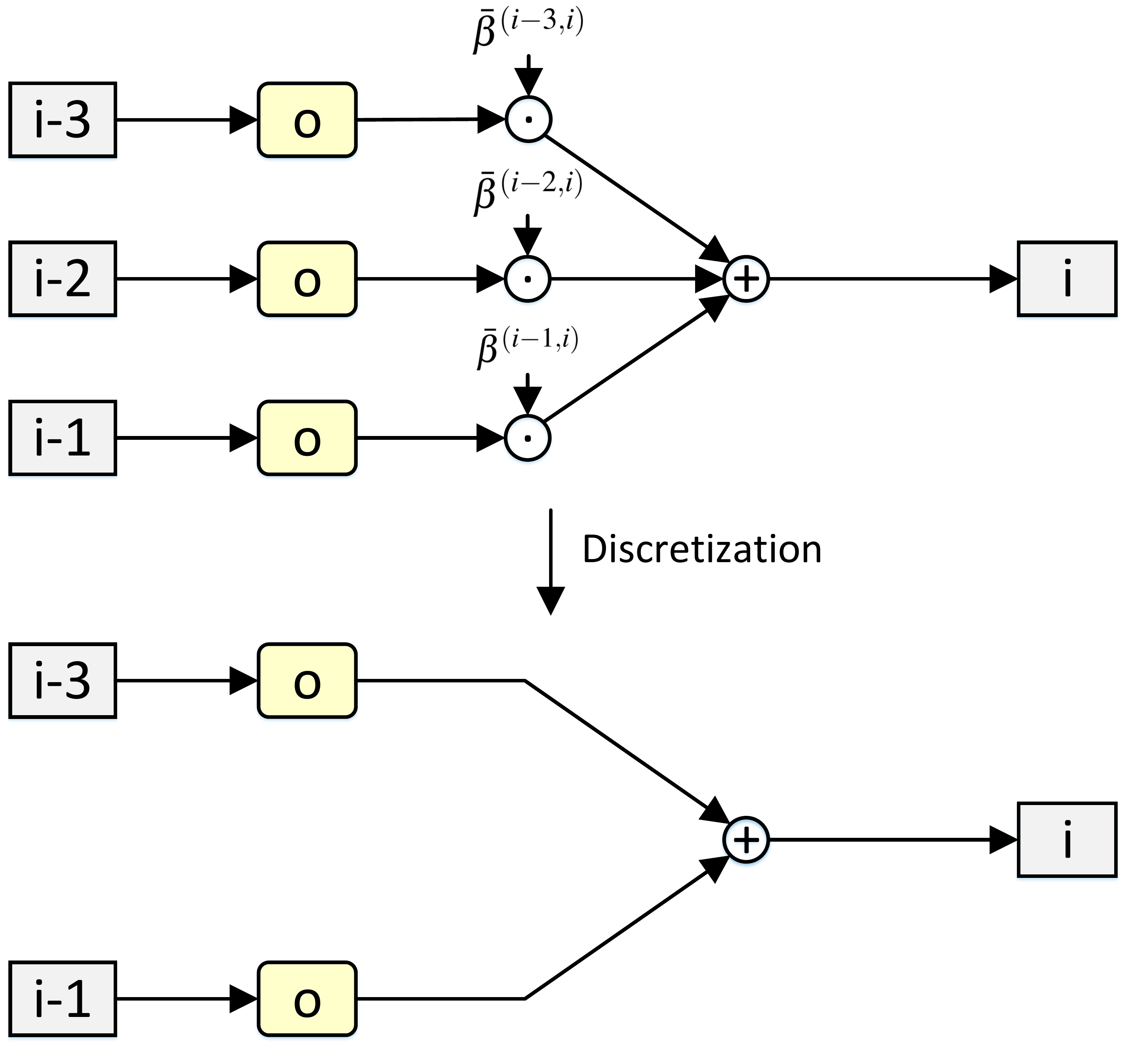}}
   \caption{\small The example of node discretization. The connections between node $i$ and predecessors with top-$k$ highest wights will be retained.}
   \label{fig:node_discrete}
   \vspace{-0.15in}
\end{figure}

Towards a natural and smooth transition from relaxed to discrete architecture, here we devise a scheduled scheme that progressively induces the optimal operation during training. Specifically, we decompose the discretization of whole architecture into several one-step discretizations performed on edges and nodes separately. For the \textbf{edge discretization}, the mixing operation in this edge will be replaced by the operation with highest weight as shown in Figure \ref{fig:edge_discrete}. Formally, when performing discretization on edge $(j, i)$, the mixing operation in Equ.~\ref{eq:out} will be switched to discrete operation
\begin{equation}
\small
\label{eq:edge_discrete}
\begin{aligned}
~~~~~~~~~~ ~~~~~~~~~~ ~~~~~~~~~~ ~~~~~ \bar o^{(j, i)}(x)=\bar\beta^{(j, i)} o^{(j, i)}(x)
\end{aligned}~~,
\end{equation}
where  $o^{(j, i)}={\arg\max}_{o \in \mathcal{O}}~\alpha_{o}^{(j, i)}$ is the optimal operation for edge $(j, i)$. After edge discretization, since the operation on this edge is resolved, we remove the weights for operations $\alpha_{o}^{(j, i)}$ but still optimize the weight $\beta^{(j, i)}$ to measure the importance of edge $(j, i)$. Besides the operation on each edge, we also gradually determine the topological structure by \textbf{node discretization}, as shown in Figure \ref{fig:node_discrete}.  Specifically, for node $i$, when the types of operations on all incident edges have been resolved, we will determine the connection between node $i$ and other nodes by only retaining the predecessors with top-$k$ strongest effect $\beta$ (e.g., $k=2$ in Figure \ref{fig:node_discrete}). After the node discretization, the computation of node $i$ will be changed from the continuous function in Equ. \ref{eq:out} to discrete function in Equ. \ref{eq:node}, and the final architecture will be obtained when all the intermediate nodes are discretized.

For each architecture discretization, we need $M=E+N$ one-step discretizations, where $E$ and $N$ is the number of edges and nodes, respectively. To control the priority and frequency of discretization, we define a scheduled variable $M_t$ as the number of one-step discretizations have been performed in the $t$-th mini-batch of the training. When $M_t=0$,  a mixture of candidate operations are initially placed on all the edges and all the predecessors are connected to all the nodes. While when $M_t$ equals the total one-step discretization number $M$, the choice of operations on all the edges are fixed as $o^{(j, i)}$ and the topological structure are also fixed. During the training procedure, we gradually discretize the edges and nodes in turn and $M_t$ increases from $0$ to $M$ as learning proceeds. With the increase of $M_t$, we select either an unresolved edge or an unresolved node with highest priority to be discretized. For each unresolved edge, we measure the priority of edge discretization by the weight difference between topmost operation and the second one. The higher weight difference indicates more confidence in the choice of operation. Analogously, for node discretization, the priority is defined as weight difference between rank-$k$ predecessor node and rank-$(k+1)$ predecessor node.

\begin{figure}[!tb]
   \centering
   \subfigure[Schedule-A.]{
     \label{fig:schedule:a}
     \includegraphics[width=0.15\textwidth]{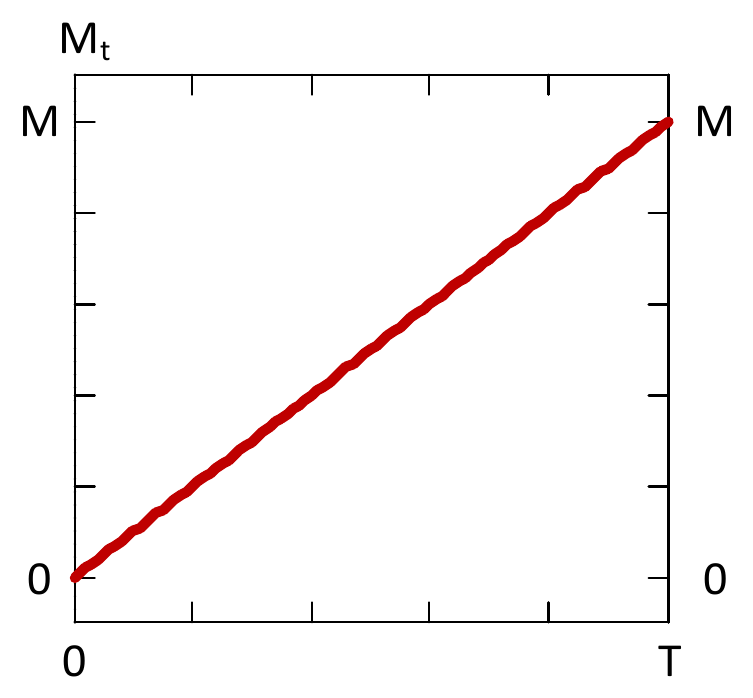}}
   \subfigure[Schedule-B.]{
     \label{fig:schedule:b}
     \includegraphics[width=0.15\textwidth]{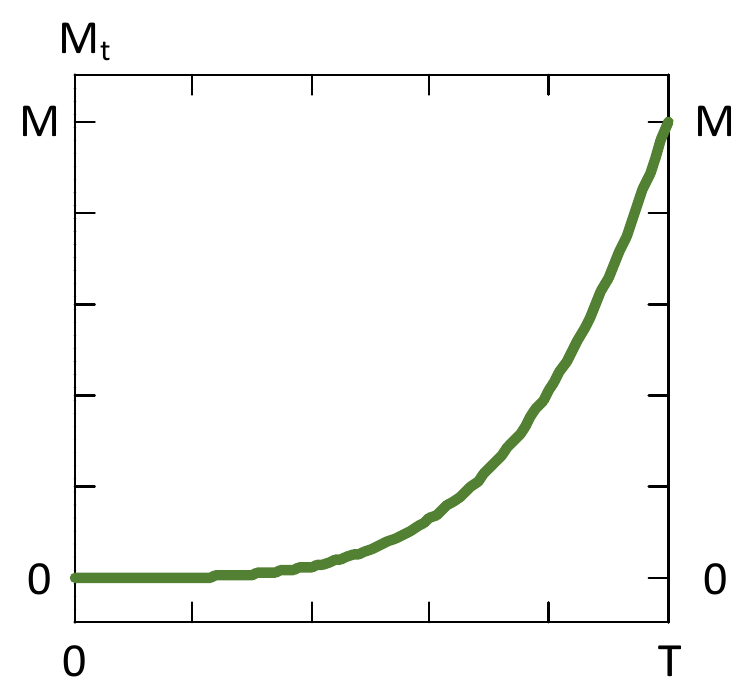}}
   \subfigure[Schedule-C.]{
     \label{fig:schedule:c}
     \includegraphics[width=0.15\textwidth]{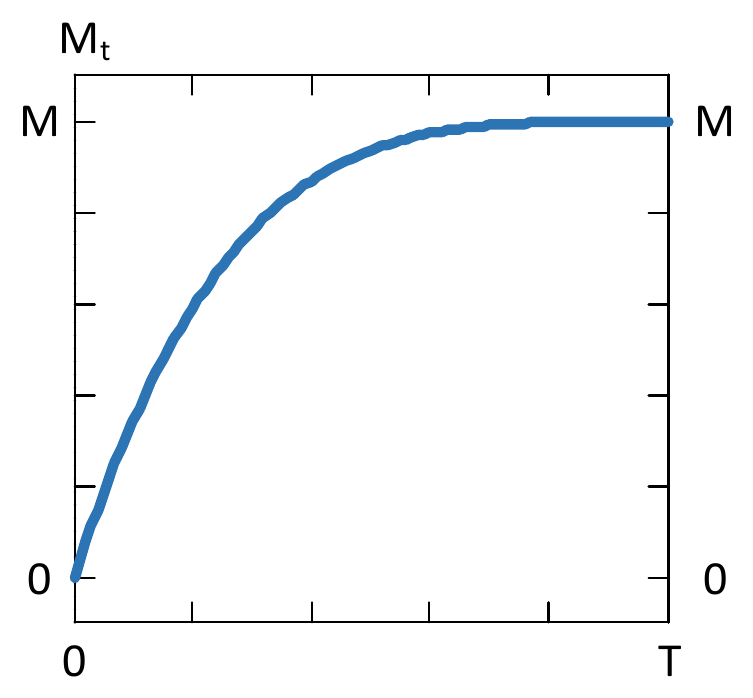}}
   \vspace{-0.10in}
   \caption{\small Examples of three schedules. }
   \label{fig:schedule}
   \vspace{-0.00in}
\end{figure}

Formally, we propose to use a schedule to increase $M_t$ as a function of $t$ itself and capitalize on such schedule to progressively perform discretization on edges and nodes. The selection of schedule also impacts the convergence speed and time cost of architecture search. Examples of three schedules are given in Figure \ref{fig:schedule} as follows:

(1) Schedule-A: $M_t=M \cdot (t/T)$ linearly increases the number of discretization decisions, where T denotes the maximum iteration.

(2) Schedule-B: $M_t=M \cdot {(t/T)}^4$ increases the number in a polynomial manner (with quartic as an example), which converges lower than linear function and discretizes the edges and nodes at the late iterations.

(3) Schedule-C: $M_t=M \cdot(1-{(1-t/T)}^4)$ utilizes a negative polynomial function. The discretization is performed at early iterations and the architecture will be converged faster than Schedule-A.

The comparisons between different schedules in SDAS will be elaborated in experiments. Algorithm \ref{alg:sdas} further details the optimization steps.

\begin{algorithm}[!tb]\small
\caption{\small Schedule DAS (SDAS)}\label{alg:sdas}
\begin{algorithmic}[1]
    \STATE \textbf{Input:}
        The set of training samples; the set of validation samples; the learning rate $\eta_1$ and $\eta_2$; the selected schedule function $M_t$; the maximum iteration $T$.
    \STATE \textbf{Initialization:}
        Randomly initialized $w$; initialize $\alpha=0, \beta=0$.
    \STATE \textbf{Output:}
        The architecture weights $\alpha, \beta$.
    \FOR{$t=1$ to $T$}
    \STATE
        Sample training mini-batch.
    \STATE
        $\triangledown w=\triangledown \mathcal{L}_{train}(w | \alpha, \beta)$.
    \STATE
        $w=w - \eta_1 \triangledown w$.
    \STATE
        Sample validation mini-batch.
    \STATE
        $\triangledown \alpha, \triangledown \beta=\triangledown \mathcal{L}_{val}(\alpha, \beta | w)$.
    \STATE
        $\alpha=\alpha - \eta_2 \triangledown \alpha$.
    \STATE
        $\beta=\beta - \eta_2 \triangledown \beta$.
    \IF{$M_t$ increases}
    \STATE
        Select the unresolved edge/node with highest priority.
    \STATE
        Perform one-step discretization.
    \ENDIF
    \ENDFOR
\end{algorithmic}
\end{algorithm}

\subsection{The View of Search Space Reduction}

In this section, we will introduce the principle of our proposed SDAS in the view of search space reduction. As discussed in Section \ref{sec:sd}, the gap between architecture in relaxed mode and architecture in discrete mode will lead to skewed optimization for DAS. One vital reason behind this gap is the enormous search space. For example, the search space for image recognition in DARTS \citep{liu2018darts} consists of $(C_{3}^{2}\times C_{4}^{2}\times C_{5}^{2}\times 7^8) ^ 2  \approx 1.07\times 10^{18}$ valid architectures. Single continuously relaxed architecture cannot well reflect the performance of this large number of discrete architectures. However, in our proposed SDAS, the search space is reduced gradually during training by discreting single edge or node. One example is shown in Table \ref{tab:searchspace}, in which we utilize SDAS with the same settings as in DARTS and the search space changes during training is given. With the variable $M_t$ increases, the search space is reduced dramatically since part of the decisions have been made during one-step discretization. The iterations after that will only consider the sub-space with valid architectures conditioned on the determined decisions. Finally, after optimization, the only valid architecture is the optimal architecture produced by SDAS.

In order to quantitatively measure the gap between architecture during search and produced architecture, Xie et al. \citep{xie2018snas} introduce to evaluate the performance of these two architectures by using the parameters $w$ learnt during search. We follow the settings in \citep{xie2018snas} and evaluate our SDAS as shown in Table \ref{tab:child}. Unsurprisingly, the discrete architecture of DARTS achieves much lower accuracy than the relaxed one, since the discretization scheme that once ignores all the other operations making the network change dramatically. By replacing the traditional softmax-relaxation by gambling softmax, the discrete architecture by SNAS can even get slightly higher performance than relaxed mode. For our SDAS, since the one-step discretization has been performed during search according to the schedule function, the relaxed architecture will converge to discrete mode which achieves the highest accuracy.

\begin{table}
\centering
\small
\caption{\small One example of search space reduction by our SDAS.}
\vspace{0.1cm}
\begin{tabular}{l|c|c|c} \hline
\begin{minipage}{3.5cm}\vspace{0.1cm}\textbf{Method} \vspace{0.1cm}\end{minipage} & \textbf{$t/T$} & \textbf{$M_t$} & \textbf{Search Space} \\ \hline
\begin{minipage}{3.5cm}\vspace{0.1cm}DARTS \citep{liu2018darts} \vspace{0.1cm}\end{minipage} & --  & -- & $1.07\times 10^{18}$ \\ \hline
\multirow{11}*{\textbf{SDAS (Schedule-A)}} & $0.0$ & $0.0$  & $1.07\times 10^{18}$ \\
& $0.1$ & $0.1$  & $9.90\times 10^{16}$ \\
& $0.2$ & $0.2$  & $4.81\times 10^{14}$ \\
& $0.3$ & $0.3$  & $4.44\times 10^{13}$ \\
& $0.4$ & $0.4$  & $2.56\times 10^{11}$ \\
& $0.5$ & $0.5$  & $2.49\times 10^{10}$ \\
& $0.6$ & $0.6$  & $1.13\times 10^{~9}$ \\
& $0.7$ & $0.7$  & $3.00\times 10^{~6}$ \\
& $0.8$ & $0.8$  & $3.19\times 10^{~4}$ \\
& $0.9$ & $0.9$  & $5.64\times 10^{~2}$ \\
& $1.0$ & $1.0$  & $1$ \\
\hline
\end{tabular}
\label{tab:searchspace}
\vspace{-0.0in}
\end{table}

\begin{table}
\centering
\small
\caption{\small Accuracy comparisons between architecture search methods by using parameters during search on CIFAR10.}
\vspace{0.1cm}
\begin{tabular}{l|c|c} \hline
\begin{minipage}{3.5cm}\vspace{0.1cm}\textbf{Method} \vspace{0.1cm}\end{minipage} & \textbf{Relaxed Mode} & \textbf{Discret Mode} \\ \hline
\begin{minipage}{3.5cm}\vspace{0.1cm}DARTS \citep{liu2018darts} \vspace{0.1cm}\end{minipage} & 87.67\% &  54.66\% \\ \hline
\begin{minipage}{3.5cm}\vspace{0.1cm}SNAS \citep{xie2018snas} \vspace{0.1cm}\end{minipage} & 88.54\%  & 90.67\% \\ \hline
\begin{minipage}{3.5cm}\vspace{0.1cm}\textbf{SDAS (Schedule-A)} \vspace{0.1cm}\end{minipage} & \multicolumn{2}{|c}{92.02\%} \\ \hline
\end{tabular}
\label{tab:child}
\vspace{-0.10in}
\end{table}

\begin{figure*}[!tb]
   \centering {\includegraphics[width=1.0\textwidth]{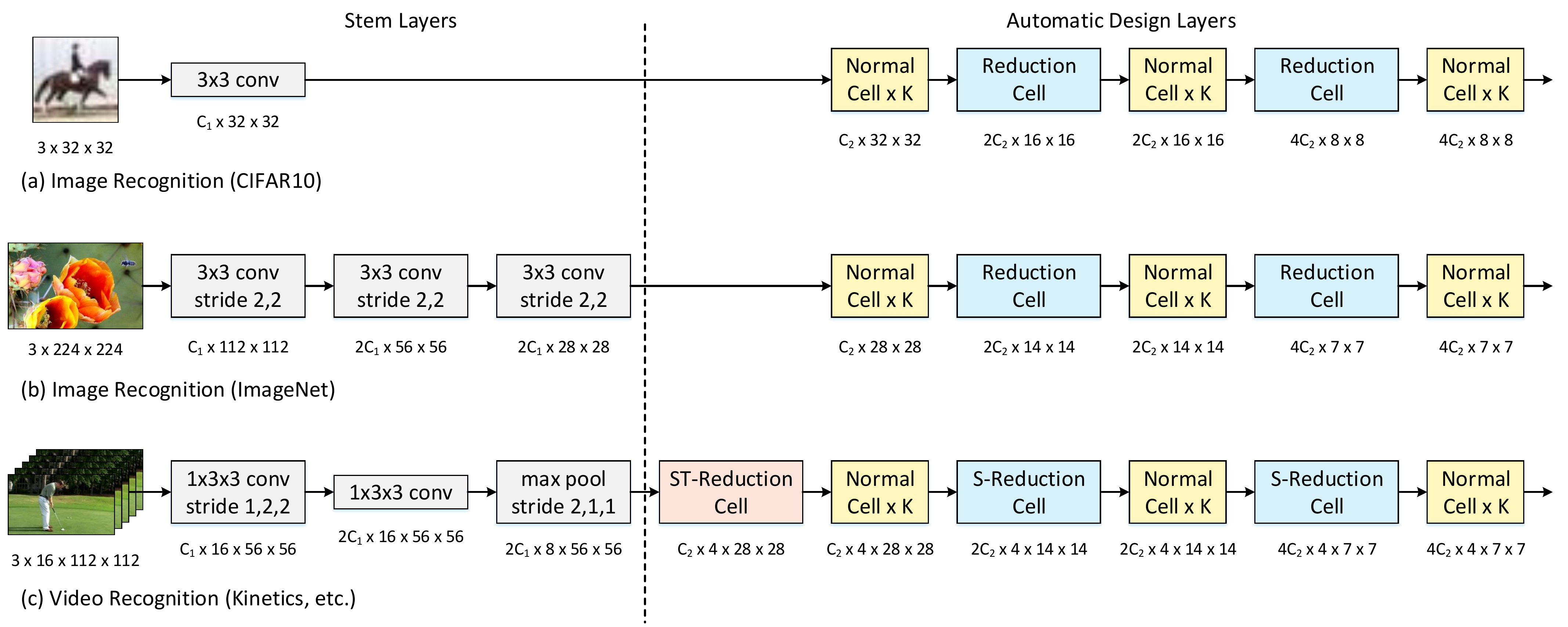}}
   \caption{\small Scalable network structures for (a) image recognition with low-resolution input (CIFAR10); (b) image recognition with high-resolution input (ImageNet); (c) video recognition (Kinetics, etc.). Each structure consists of stem layers and automatic design layers, where the stem layers are manually determined. The automatic design layers are composed of a stack of cells and each cell is automatically designed by SDAS.}
   \label{fig:structure}
   \vspace{-0.15in}
\end{figure*}

\section{Network Structure for Visual Recognition}

\subsection{Scalable Network Structure}
Figure \ref{fig:structure} shows an overview of the scalable network structures for (a) image recognition with low-resolution input; (b) image recognition with high-resolution input; (c) video recognition. Each network mainly consists of two types of layers, i.e., stem layers and automatic design layers. The stem layers are manually determined to learn the low-level representations from the input data. The main purpose of stem layers is to reduce the spatial or temporal dimensions, so as to decrease computational cost and GPU memory demand. The automatic design layers are composed of a stack of computation architectures or cells. Each cell is automatically designed by our SDAS and has the same structure, but different weights. For image recognition, we define two kinds of computation cells, each with different property. \textbf{Normal Cell} preserves the resolution of the inputs.  \textbf{Reduction Cell} produces the output feature map with the spatial resolution reduced by a factor of two. While for video recognition, we further divide the Reduction Cell into \textbf{S-Reduction Cell} and \textbf{ST-Reduction Cell} for reducing only spatial dimension and reducing both spatio-temporal dimension, respectively. For all the Reduction Cells, we make the initial operation applied on the input nodes with the corresponding stride to change the resolution or duration. The architecture parameters $\alpha, \beta$ is the combination of parameters for each cell, which are optimized together during search process. The placement of different cells is detailed in the Figure \ref{fig:structure}, in which $K$ is the repeat number of Normal Cell, $C_1$ and $C_2$ denotes the number of output channels in the stem layers and the first cell, respectively. The three numbers are considered as free parameters to make the network structure tailored to the scale of visual recognition problem.

\subsection{Operation Set}

Inspired by the recent advances in CNN, we start from an operation set $\mathcal{O}_{2D}$, which consists of several 2D operations. All of them are prevalent in image recognition \citep{liu2018darts,xie2018snas}:

\begin{table}[H]
    \footnotesize
    \setlength{\tabcolsep}{0.5em} 
    \centering
    \begin{tabular}{l@{~~~~~~~~~~~~}l}
    $\bullet\;$ identity                              & $\bullet\;$ 3$\times$3 ave pooling   \\
     $\bullet\;$ 3$\times$3 max pooling                     & $\bullet\;$ 3$\times$3 separable conv \\
    $\bullet\;$ 5$\times$5 separable conv                  & $\bullet\;$ 3$\times$3 dilated separable conv \\
    $\bullet\;$ 5$\times$5 dilated separable conv
    \end{tabular}
    \vspace{-0.1in}
\end{table}

The $\mathcal{O}_{2D}$ includes three types of operations, i.e., identity shortcut, 2D pooling and 2D convolution. For each pooling or convolution, we denote the local window size as $k_s \times k_s$ where $k_s$ is the spatial size. The separable convolution \citep{Chollet2017CVPR} factorizes the standard convolution into a depth-wise convolution and a point-wise convolution for a good tradeoff between computation cost and performance, and is always applied twice in an operation \citep{liu2018darts,zoph2018learning}. The dilated convolution \citep{yu2016multi} further enlarges the receptive field of each convolution by atrous rate 2. For each convolutional operation, we exploit ReLU-Conv-BN order.

When utilize the operation set $\mathcal{O}_{2D}$ to video recognition, we extend the local window size of the operation to $k_t \times k_s \times k_s$, where $k_t$ is the temporal duration of local window. Here, we consider the operations with $k_t=1$ as 2D operations since they are performed on each frame independently. Therefore, the $\mathcal{O}_{2D}$ for video recognition is given as

\begin{table}[H]
    \footnotesize
    \setlength{\tabcolsep}{0.5em} 
    \centering
    \begin{tabular}{l@{~~~~~~}l}
    $\bullet\;$ identity                              & $\bullet\;$ 1$\times$3$\times$3 ave pooling   \\
     $\bullet\;$ 1$\times$3$\times$3 max pooling                     & $\bullet\;$ 1$\times$3$\times$3 separable conv \\
    $\bullet\;$ 1$\times$5$\times$5 separable conv                  & $\bullet\;$ 1$\times$3$\times$3 dilated separable conv \\
    $\bullet\;$ 1$\times$5$\times$5 dilated separable conv
    \end{tabular}
    \vspace{-0.1in}
\end{table}

To fully explore the temporal evolution across consecutive frames, we build the following set $\mathcal{O}_{3D}$ by extending the operations in $\mathcal{O}_{2D}$ to 3D manner:

\begin{table}[H]
    \footnotesize
    \setlength{\tabcolsep}{0.5em} 
    \centering
    \begin{tabular}{l@{~~}l}
    $\bullet\;$ identity                              & $\bullet\;$ 1$\times$3$\times$3 max pooling   \\
    $\bullet\;$ 1$\times$3$\times$3 separable conv & $\bullet\;$ 1$\times$5$\times$5 separable conv\\
    $\bullet\;$ 1$\times$3$\times$3 dilated separable conv & $\bullet\;$ 3$\times$3$\times$3 max pooling \\
    $\bullet\;$ 3$\times$3$\times$3 separable-3d conv                  & $\bullet\;$ 3$\times$5$\times$5 separable-3d conv \\
    $\bullet\;$ 3$\times$3$\times$3 dilated separable-3d conv \\
    \end{tabular}
    \vspace{-0.1in}
\end{table}

\begin{figure}[!tb]
   \centering
   \subfigure[Separable-3d convolution.]{
     \label{fig:ops:a}
     \includegraphics[width=0.42\textwidth]{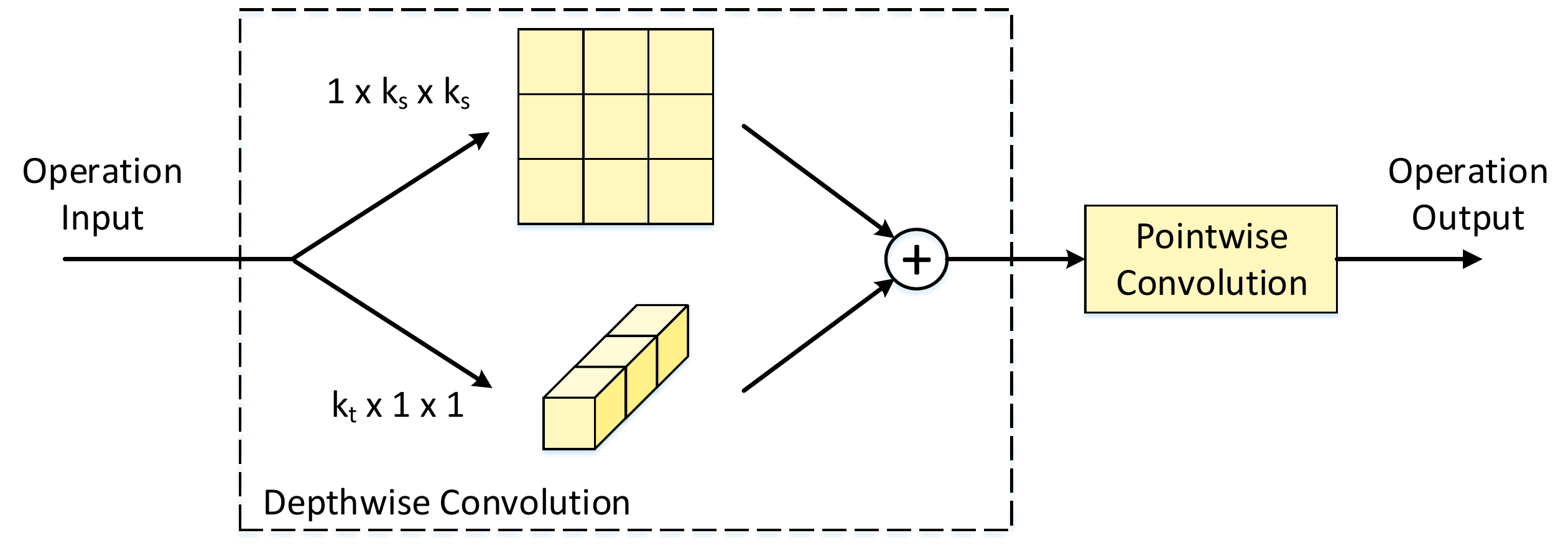}}
   \subfigure[Channel-wise Scale operation.]{
     \label{fig:ops:b}
     \includegraphics[width=0.42\textwidth]{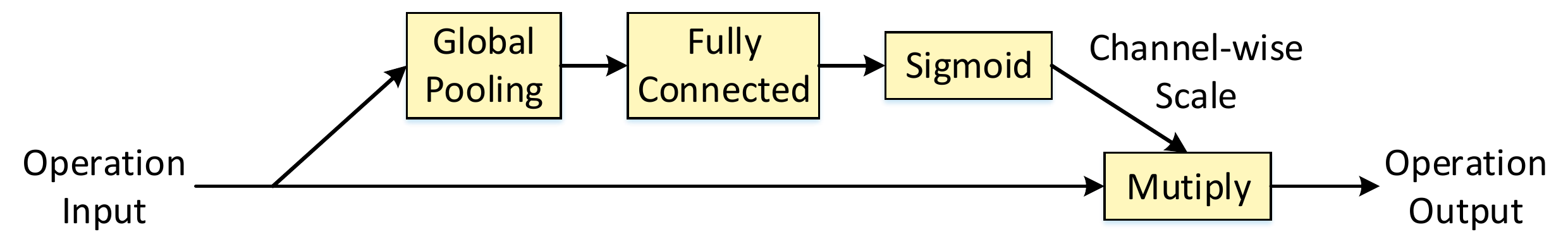}}
   \subfigure[Channel-wise Bias operation.]{
     \label{fig:ops:c}
     \includegraphics[width=0.42\textwidth]{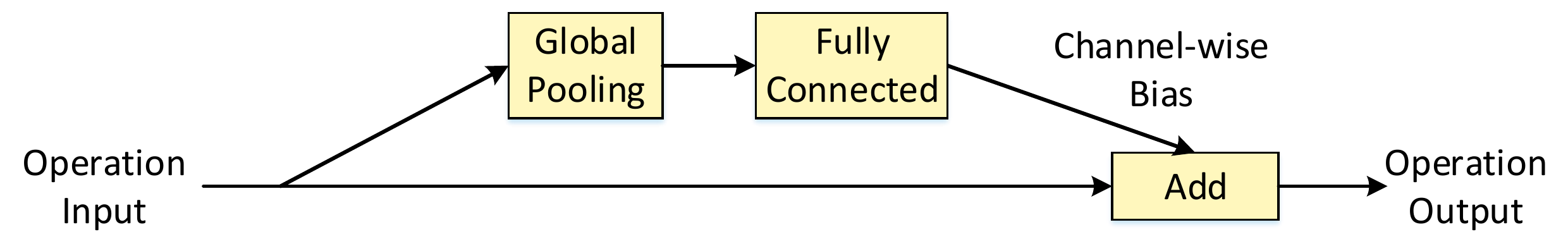}}
   \caption{\small The schematic illustration of Separable-3d convolution, Channel-wise Scale operation and Channel-wise Bias operation.}
   \label{fig:ops}
   \vspace{-0.15in}
\end{figure}

Here, we remove the most infrequent operations, i.e., average pooling and $5\times 5$ dilated convolution to control the size of operation set. One unique design is \textbf{Separable-3d convolution}, in which we remould the idea in \citep{qiu2017learning} of decomposing 3D learning into 2D convolution in spatial space and 1D operation in temporal dimension. The structure is shown in Figure \ref{fig:ops:a}. Separable-3d convolution simulates each $k_t\times k_s \times k_s$ convolution with one $1\times k_s \times k_s$ depthwise convolution plus one $k_t\times 1 \times 1$ depthwise convolution. The outputs from the two are accumulated and input into pointwise convolution to obtain the final output. Such convolution is conceptually similar to P3D-B block as presented in \citep{qiu2017learning}, which places spatial convolution and temporal convolution in parallel.

In addition to the operation set $\mathcal{O}_{3D}$, we also construct an advanced operation set $\mathcal{O}_{adv}$ which adds two new channel-wise operations to $\mathcal{O}_{3D}$. The new operations are derived from the idea in Squeeze-and-Excitation Networks (SENet) \citep{hu2018squeeze}. The goal of SENet is to improve the representational power of a network by explicitly modeling the interdependencies between the channels of feature map. As a result, we summarize the key component of SENet as two standalone operations, namely \textbf{Channel-wise Scale operation} and \textbf{Channel-wise Bias operation}, which recalibrates channel-wise feature responses by adaptive scale and bias, respectively. Figure \ref{fig:ops:b} details the Channel-wise Scale operation, in which the channel-wise weights are achieved by the global-average-pooled representation, and then utilized to re-scale the input feature map. For Channel-wise Bias operation in Figure \ref{fig:ops:c}, the adaptive bias for each channel is learnt instead and broad-casted to each position in the input feature map. By the proposal of these two operations, the advanced operation set $\mathcal{O}_{adv}$ is defined as

\begin{table}[H]
    \footnotesize
    \setlength{\tabcolsep}{0.5em} 
    \centering
    \begin{tabular}{l@{~~~~~~~~}l}
    $\bullet\;$ identity                              & $\bullet\;$ 1$\times$3$\times$3 max pooling   \\
    $\bullet\;$ 1$\times$3$\times$3 separable conv & $\bullet\;$ 1$\times$5$\times$5 separable conv\\
    $\bullet\;$ 3$\times$3$\times$3 max pooling &  $\bullet\;$ 3$\times$3$\times$3 separable-3d conv\\
    $\bullet\;$ 3$\times$5$\times$5 separable-3d conv & $\bullet\;$ channel-wise scale \\
    $\bullet\;$ channel-wise bias
    \\
    \end{tabular}
    \vspace{-0.1in}
\end{table}

Similarly, the infrequent operation $3\times 3$ dilated convolution is removed for a more exquisite operation set. The comparison between different operation sets will be discussed in the experiments.

\section{Implementation}
\subsection{Datasets}

We empirically evaluate our SDAS on CIFAR10 and ImageNet dataset for image recognition. The \textbf{CIFAR10} \citep{krizhevsky2009learning} dataset consists of 60K images with low-resolution ($32\times 32$) in 10 classes, and \textbf{ImageNet} \citep{russakovsky2015imagenet} consists of around 1.2M image in 1K categories. We use the official training/test split provided by the dataset organizers. Following the settings in \citep{liu2018darts}, we search the optimal architecture on CIFAR10 dataset, and then utilize the produced architecture on both CIFAR10 and ImageNet dataset.

For video recognition, we conduct the experiments on three public benchmarks, i.e., UCF101, HMDB51, and Kinetics400. \textbf{UCF101} \citep{UCF101} and \textbf{HMDB51} \citep{HMDB51} are two of the most popular video action recognition benchmarks. UCF101 consists of 13K videos from 101 action categories, and HMDB51 consists of 7K videos from 51 action categories. Each split in UCF101 includes about 9.5K training and 3.7K test videos, while a HMDB51 split contains 3.5K training and 1.5K test videos. The \textbf{Kinetics400} \citep{carreira2017quo} dataset is one of the large-scale action recognition benchmarks. It consists of around 300K videos from 400 action categories. The 300K videos are divided into 240K, 20K, 40K for training, validation and test set, respectively. Each video in this dataset is 10-second short clip cropped from the raw YouTube video. Note that the labels for test set are not publicly available and the performances on Kinetics400 dataset are all reported on the validation set. In addition, we conduct a subset of Kinetics dataset, called \textbf{Kinetics10}, which consists of 10 categories, by merging the similar fine-grained categories. The details about categories in Kinetics10 dataset is given in Table \ref{tab:k10c}. Such subset contains 40K training videos and 3K validation videos. Similar to image recognition, we search the optimal architecture on UCF101, HMDB51 and Kinetics10 dataset, and then evaluate on the large-scale Kinetics400 dataset.

\begin{table}
\centering
\small
\caption{\small The categories in Kinetics10 dataset.}
\vspace{0.1cm}
\begin{tabular}{l|c} \hline
\textbf{Category} & \textbf{Subcategories} \\ \hline
\textbf{basketball} & \begin{minipage}{5.5cm}\vspace{0.12cm}dribbling basketball, dunking basketball, playing basketball, shooting basketball\vspace{0.12cm}\end{minipage} \\ \hline
\textbf{biking} & \begin{minipage}{5.5cm}\vspace{0.12cm}biking through snow, falling off bike, jumping bicycle, riding a bike\vspace{0.12cm}\end{minipage} \\ \hline
\textbf{bodybuilding} & \begin{minipage}{5.5cm}\vspace{0.12cm}clean and jerk, deadlifting, pull ups, situp, snatch weight lifting, squat, stretching arm, stretching leg\vspace{0.12cm}\end{minipage} \\ \hline
\textbf{cooking} & \begin{minipage}{5.5cm}\vspace{0.12cm}chopping vegetables, cooking egg, cooking on campfire, cooking sausages (not on barbeque), cooking scallops, flipping pancake, frying vegetables, making a sandwich, scrambling eggs\vspace{0.12cm}\end{minipage} \\ \hline
\textbf{dancing} & \begin{minipage}{5.5cm}\vspace{0.12cm}belly dancing, breakdancing, country line dancing, dancing ballet, dancing charleston, dancing gangnam style, dancing macarena, jumping into pool, mosh pit dancing, salsa dancing, square dancing, swing dancing, tango dancing, tap dancing \vspace{0.12cm}\end{minipage} \\ \hline
\textbf{eating} & \begin{minipage}{5.5cm}\vspace{0.12cm}eating burger, eating cake, eating carrots, eating chips, eating doughnuts, eating hotdog, eating ice cream, eating spaghetti, eating watermelon\vspace{0.12cm}\end{minipage} \\ \hline
\textbf{golfing} & \begin{minipage}{5.5cm}\vspace{0.12cm}golf chipping, golf driving, golf putting\vspace{0.12cm}\end{minipage} \\ \hline
\textbf{skiing} & \begin{minipage}{5.5cm}\vspace{0.12cm}ski jumping, skiing crosscountry, skiing mono, skiing slalom, snowboarding, tobogganing\vspace{0.12cm}\end{minipage} \\ \hline
\textbf{soccer} & \begin{minipage}{5.5cm}\vspace{0.12cm}juggling soccer ball, kicking soccer ball, passing soccer ball, shooting goal (soccer) \vspace{0.12cm}\end{minipage} \\ \hline
\textbf{swimming} & \begin{minipage}{5.5cm}\vspace{0.12cm}ice swimming, swimming backstroke, swimming breast stroke, swimming butterfly stroke, swimming front crawl \vspace{0.12cm}\end{minipage} \\ \hline
\end{tabular}
\label{tab:k10c}
   \vspace{-0.15in}
\end{table}

\subsection{Data Preprocessing.}
During training on CIFAR10 and ImageNet, the dimension of input image is set as $32 \times 32$ and $224 \times 224$ which is randomly cropped from the padded/resized image with short edge 40/256, respectively. For video recognition, the input video clip is $16\times 112 \times 112$ volume cropped from the non-overlapping 16-frame clip with short edge 128. Each input is randomly flipped along horizontal direction for data augmentation. At the inference time, we resize the input data according to shorter side, and perform inference on the center crop. Thus, the network produces one score for each image/clip, and the video-level prediction score is calculated by averaging all scores from 20 uniformly sampled clips.

\subsection{Architecture Search.}
For architecture search, we randomly split the original training set into two equal parts as training and validation sets to optimize the network architecture. Note that the original validation set is never utilized in the optimization of architecture search.
Our proposed SDAS is implemented on Caffe \citep{jia2014caffe} platform and mini-batch Stochastic Gradient Descent (SGD) algorithm is exploited to optimize the model. We set each mini-batch as 64 images/clips, which are implemented with multiple NVidia Titan Xp GPUs in parallel. For the parameter $w$, we use the momentum SGD with initial learning rate $\eta_1=0.025$ which is annealed down to zero following a cosine decay, while the weights of architecture $\alpha, \beta$ are optimized by Adam algorithm with fixed learning rate $\eta_2=3\times 10^{-4}$. The search completes after 50/320/320/96 epochs for CIFAR10, UCF101, HMDB51 and Kinetics10, respectively.

\section{Experiments on Image Recognition}

\subsection{Evaluations on CIFAR10}
In architecture search on CIFAR10, we utilize the operation set $\mathcal{O}_{2D}$, and each cell in our experiments consists of $N=6$ nodes, which include 2 input nodes plus 4 intermediate nodes. Please note that the search space in this setting is the same as in \citep{liu2018darts,xie2018snas}. During search, we set the repeat number $K$ of Normal Cell as 2, and the output channels are fixed as $C_1=48$ and $C_2=64$. Once the architecture of each cell is optimized, we increase the learning capacity of network by using $K=6$, $C_1=108$ and $C_2=144$ in the architecture evaluation. We re-train the searched architectures on original CIFAR10 training set for 600 epochs with batch size 96. Additional enhancements in \citep{liu2018darts}, including cutout, path dropout and auxiliary towers, are all exploited. As such, DAS in this paper is exactly the same setting as the DARTS (first order), which is the very comparable baseline to our SDAS.

\begin{figure*}[!tb]
   \centering
   \subfigure[0 epoch, $1.03\times 10^9$ valid architectures.]{
     \label{fig:procedure:a}
     \includegraphics[width=0.32\textwidth]{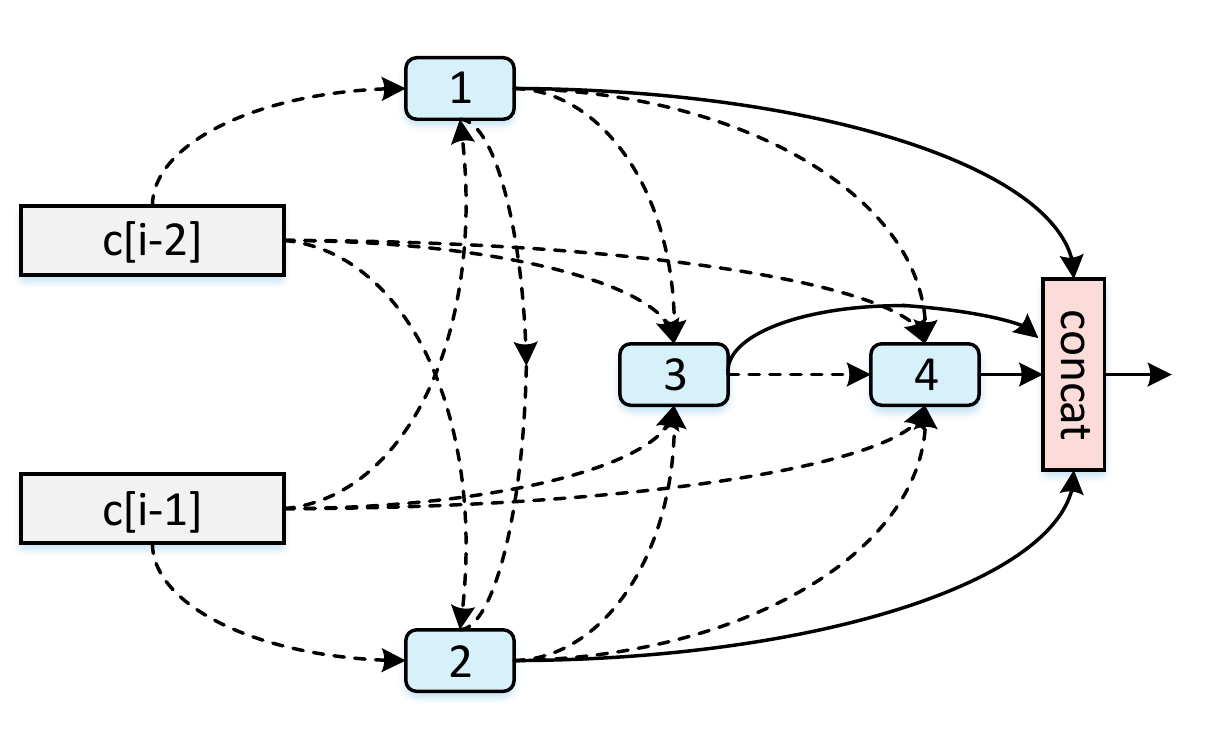}}
   \subfigure[10 epochs, $5.68\times 10^6$ valid architectures.]{
     \label{fig:procedure:b}
     \includegraphics[width=0.32\textwidth]{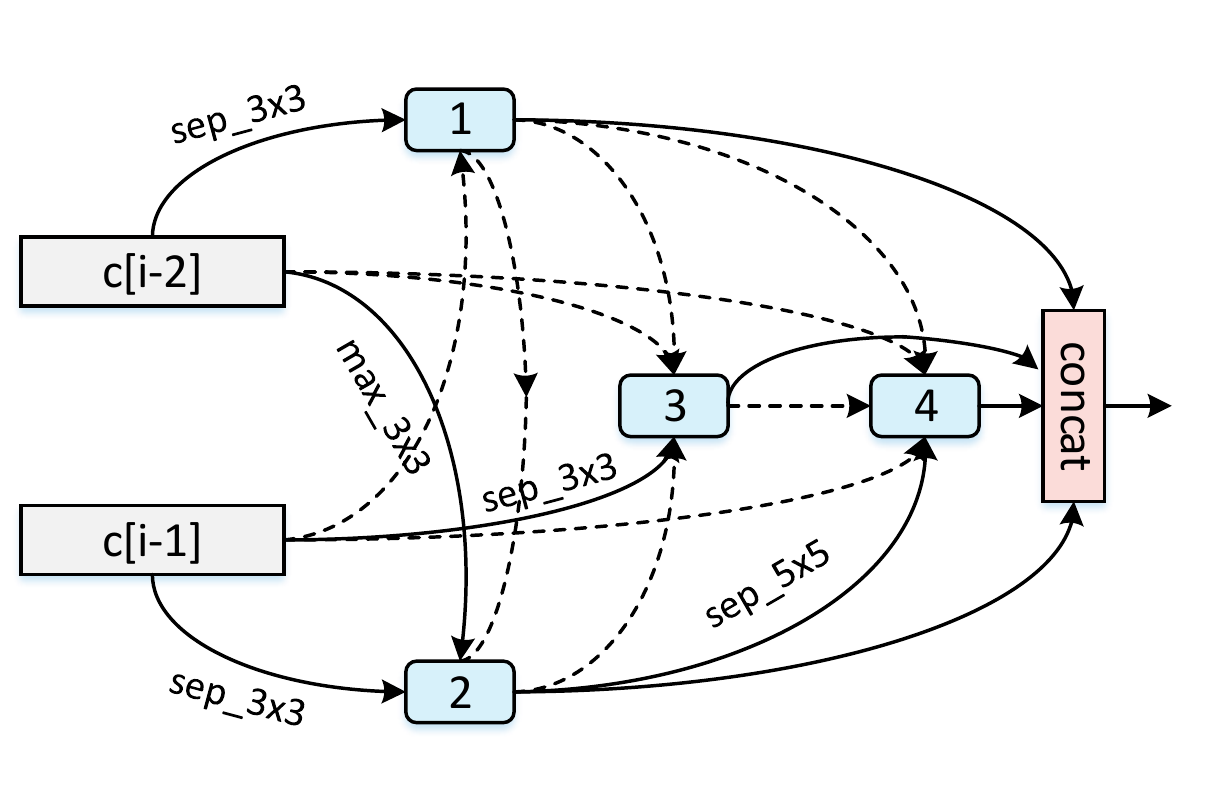}}
   \subfigure[20 epochs, $2.31\times 10^4$ valid architectures.]{
     \label{fig:procedure:c}
     \includegraphics[width=0.32\textwidth]{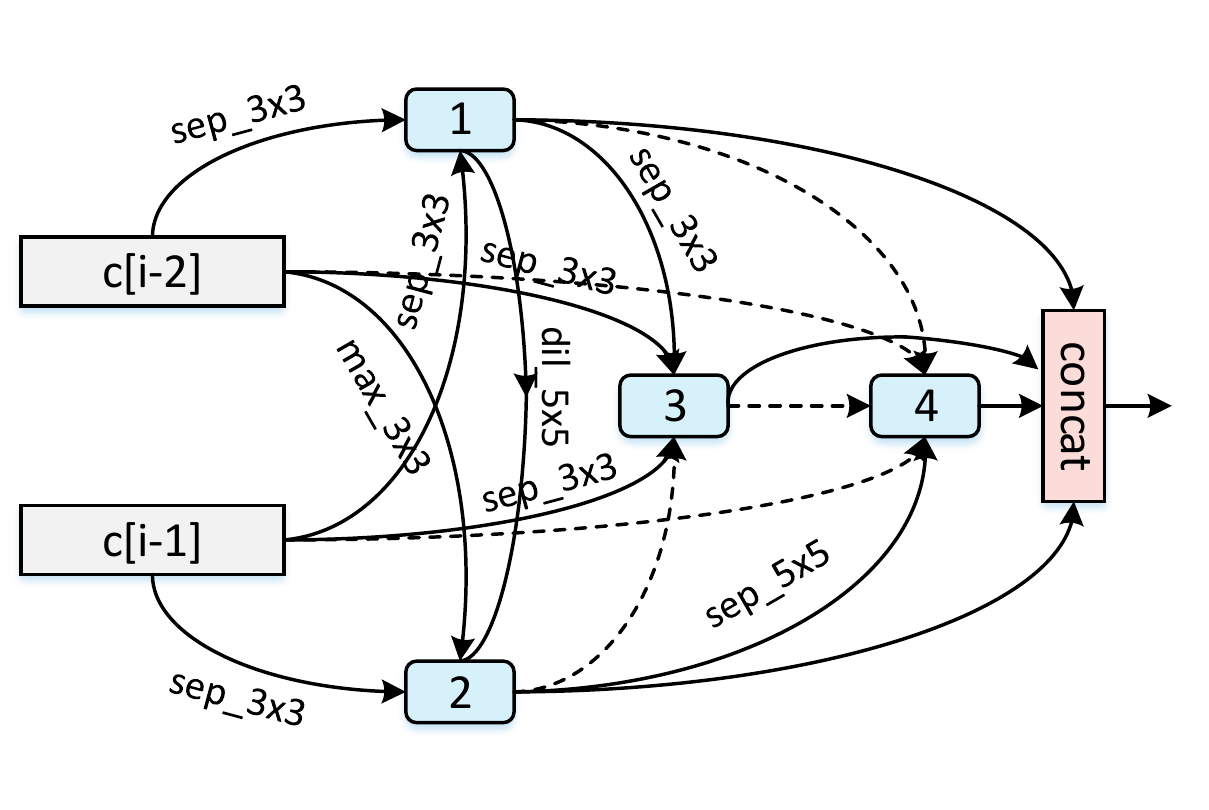}}
    \subfigure[30 epochs, $4.56\times 10^3$ valid architectures.]{
     \label{fig:procedure:d}
     \includegraphics[width=0.32\textwidth]{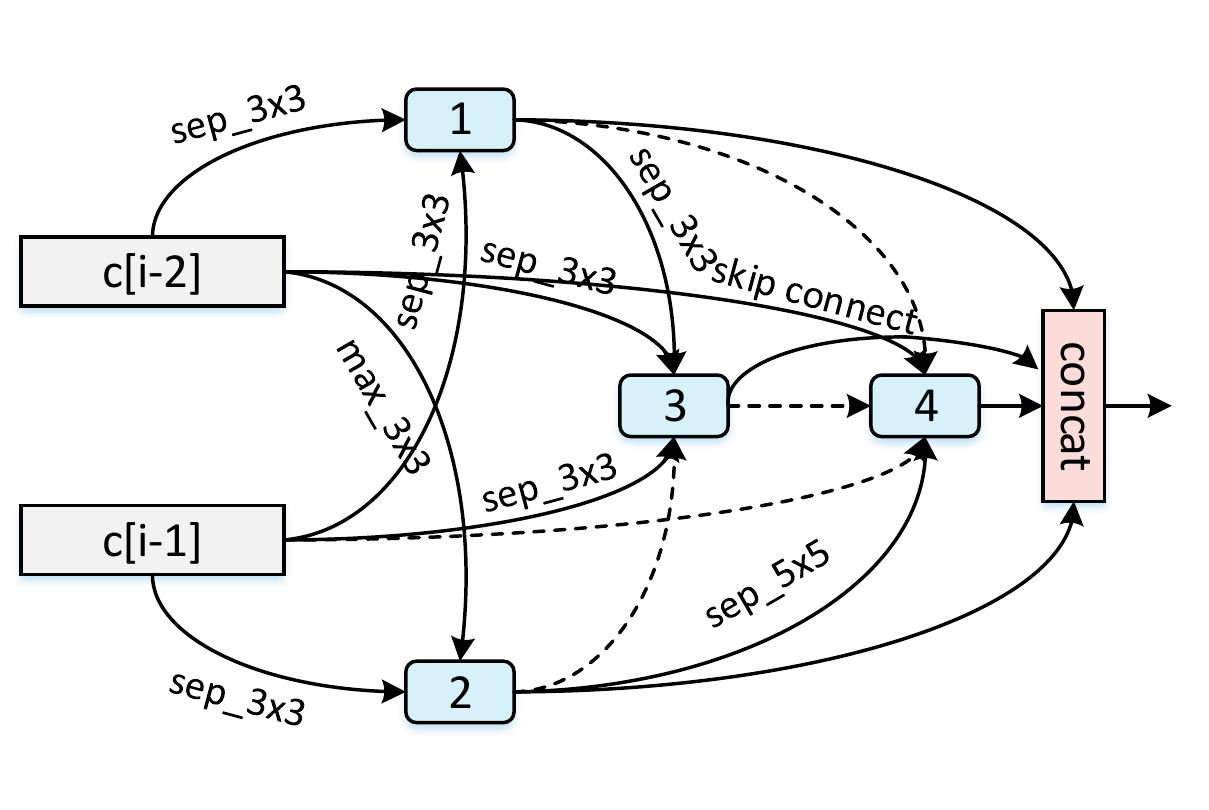}}
   \subfigure[40 epochs, $4.56\times 10^3$ valid architectures.]{
     \label{fig:procedure:e}
     \includegraphics[width=0.32\textwidth]{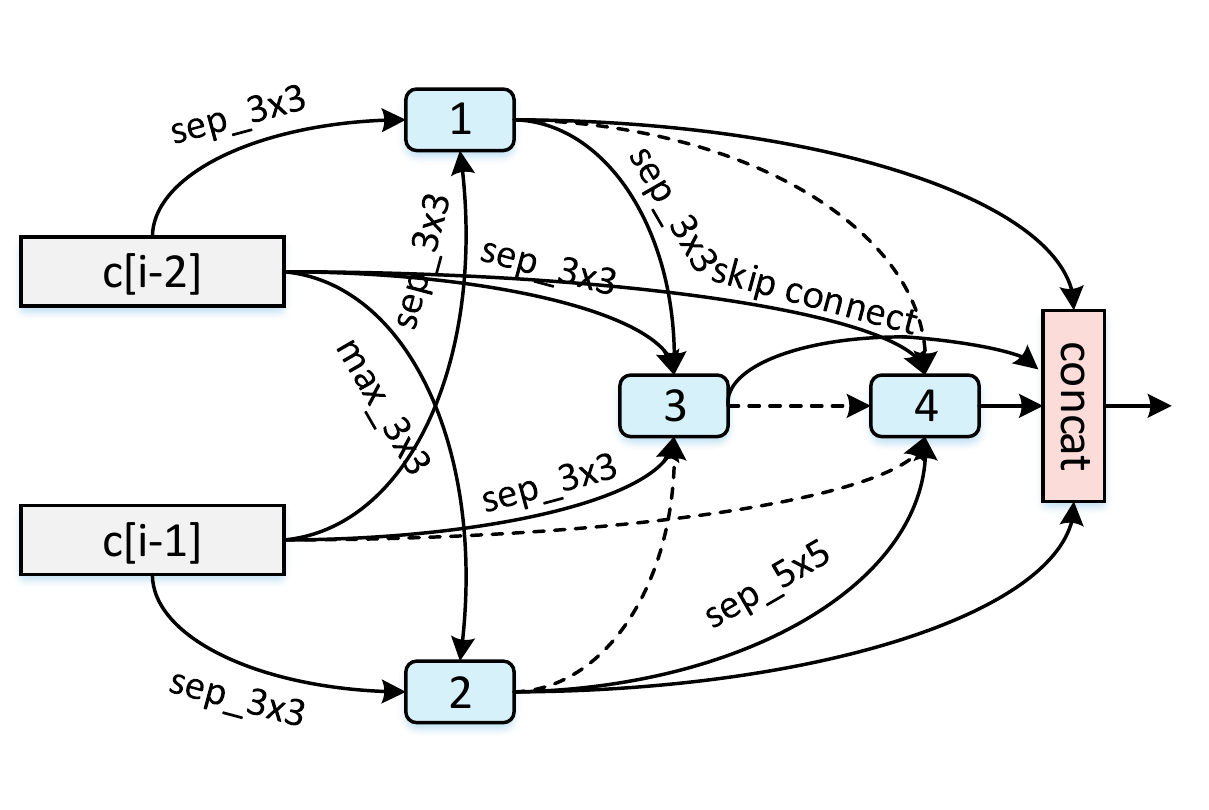}}
   \subfigure[50 epochs, $1$ valid architecture.]{
     \label{fig:procedure:f}
     \includegraphics[width=0.32\textwidth]{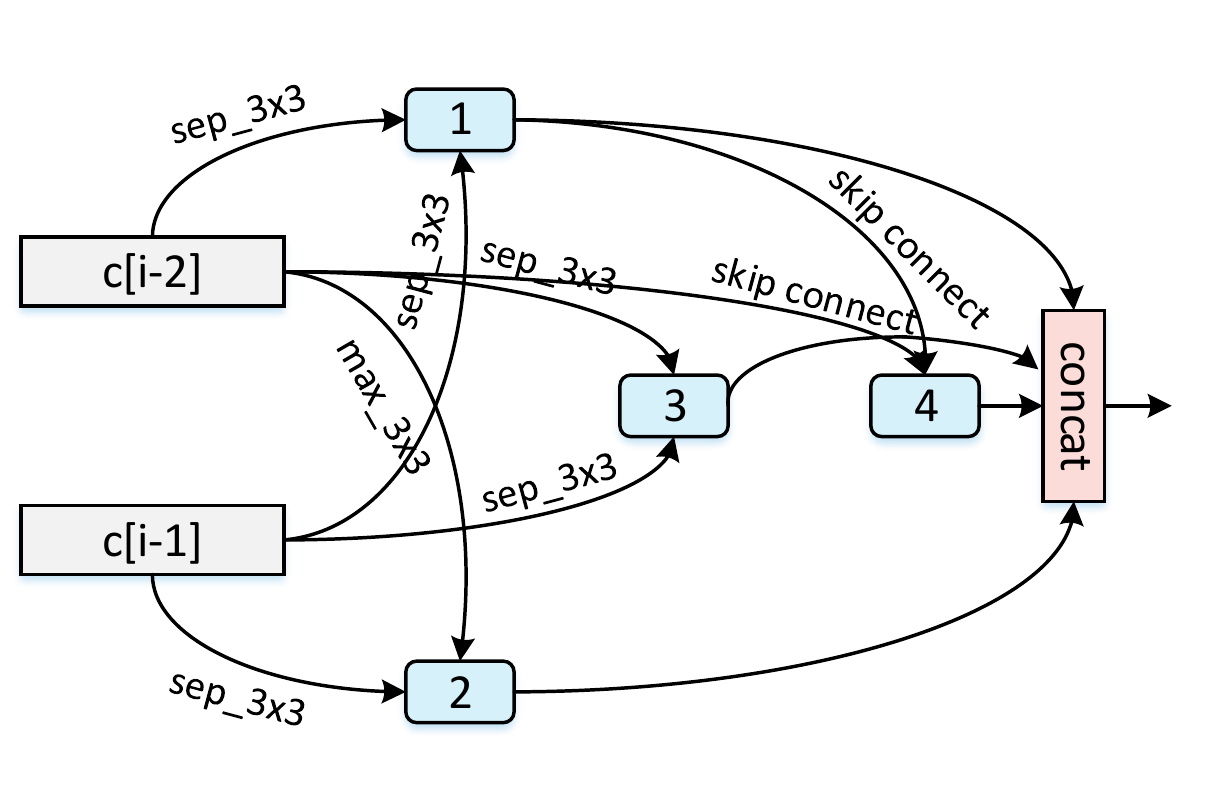}}
     \vspace{-0.10in}
   \caption{\small Examples of the Normal Cell during SDAS procedure on CIFAR10 dataset. The input nodes (gray) are the output of two prior cells. The output (pink) is the result of a concatenation operation across all resulting branches. The dash line means the operation on this edge is not determined. Please not that the Norm Cell keep the same during 30 epochs and 40 epochs, since the one-step discretization at that time is only performed on Reduction Cell.}
   \label{fig:procedure}
   \vspace{-0.10in}
\end{figure*}

\begin{table*}[]
\centering
\small
\caption{Comparisons with state-of-the-art image classifiers on CIFAR10. $^*$ denotes the search cost that is reported in the original DARTS paper \citep{liu2018darts} but with different GPU from ours. In our implementations, the time cost of DARTS (1st order) and DARTS (2nd order) is 0.5 and 1.9 GPU days, respectively.}
\begin{tabular}{@{}lccccc@{}} \hline
\multirow{2}{*}{\textbf{Architecture}}                    & \textbf{Test Error} & \textbf{Params} & \textbf{Search Cost} & \textbf{Search} \\
                                                          & \textbf{(\%)}       & \textbf{(M)}    & \textbf{(GPU days)}  & \textbf{Method} \\ \hline
 DenseNet-BC \citep{huang2017densely}                      & 3.46                & 25.6            & --                   & manual \\
 NASNet-A + cutout \citep{zoph2018learning}                & 2.65                & 3.3             & 1800                 & RL \\ 
 AmoebaNet-A + cutout \citep{real2018regularized}          & 3.34 $\pm$ 0.06     & 3.2             & 3150                 & evolution \\ 
 AmoebaNet-B + cutout \citep{real2018regularized}          & 2.55 $\pm$ 0.05     & 2.8             & 3150                 & evolution \\ 
 Hierarchical Evo \citep{liu2018hierarchical}              & 3.75 $\pm$ 0.12     & 15.7            & 300                  & evolution \\ 
 PNAS \citep{liu2018progressive}                           & 3.41 $\pm$ 0.09     & 3.2             & 225                  & SMBO \\
 ENAS + cutout \citep{pham2018efficient}                   & 2.89                & 4.6             & 0.5                  & RL \\ \hline
 DARTS (first order) + cutout \citep{liu2018darts}         & 2.94                & 2.9             & 1.5 $^*$             & gradient-based  \\
 DARTS (second order) + cutout \citep{liu2018darts}        & 2.83 $\pm$ 0.06     & 3.4             & 4 $^*$               & gradient-based  \\
 SNAS + mild + cutout      \citep{xie2018snas}             & 2.98                & 2.9             & 1.5                  & gradient-based  \\ 
 SNAS + moderate + cutout  \citep{xie2018snas}             & 2.85 $\pm$ 0.02     & 2.8             & 1.5                  & gradient-based  \\ 
 SNAS + aggressive + cutout\citep{xie2018snas}             & 3.10 $\pm$ 0.04     & 2.3             & 1.5                  & gradient-based  \\ \hline 
 Random + cutout                                          & 3.49                & 3.1             & --                   & --       \\
 DAS + cutout                                             & 2.93 $\pm$ 0.08     & 2.7             & 0.52                  & gradient-based \\
 SDAS-A + cutout                                  & 2.81 $\pm$ 0.07     & 2.9             & 0.31                  & gradient-based  \\
 SDAS-B + cutout                                  & 2.74 $\pm$ 0.11     & 3.2             & 0.37                  & gradient-based  \\
 SDAS-C + cutout                                  & 2.84 $\pm$ 0.12     & 3.9             & 0.25                  & gradient-based  \\
  \hline
\end{tabular}
\vspace{-0.10in}
\label{tab:cifar10}
\end{table*}

Figure \ref{fig:procedure} illustrates the architecture evolution of a Normal Cell during SDAS optimization with Schedule-A. Let us look at how the one-step discretization is performed on Normal Cell during SDAS optimization, which gradually achieves the optimal discrete architecture starting from relaxed mode. As shown in Figure \ref{fig:procedure:b} and Figure \ref{fig:procedure:c}, at the beginning of optimization, one-step discretization will be utilized on the most confident edge, and replace mixed operation with the optimal operation on this edge. While in Figure \ref{fig:procedure:d}, since all the incident edge of node 2 have been discretized, the SDAS considered to discretize this node. Thus, the connection between node 1 and node 2 is removed by node discretization on node 2. When reaching the maximum epochs in Figure \ref{fig:procedure:f}, all the edges and nodes are discretized and search space only contains the optimal architecture.

\begin{figure*}[!tb]
   \centering
   \subfigure[Normal Cell by SDAS-A.]{
     \label{fig:cifar:a}
     \includegraphics[width=0.32\textwidth]{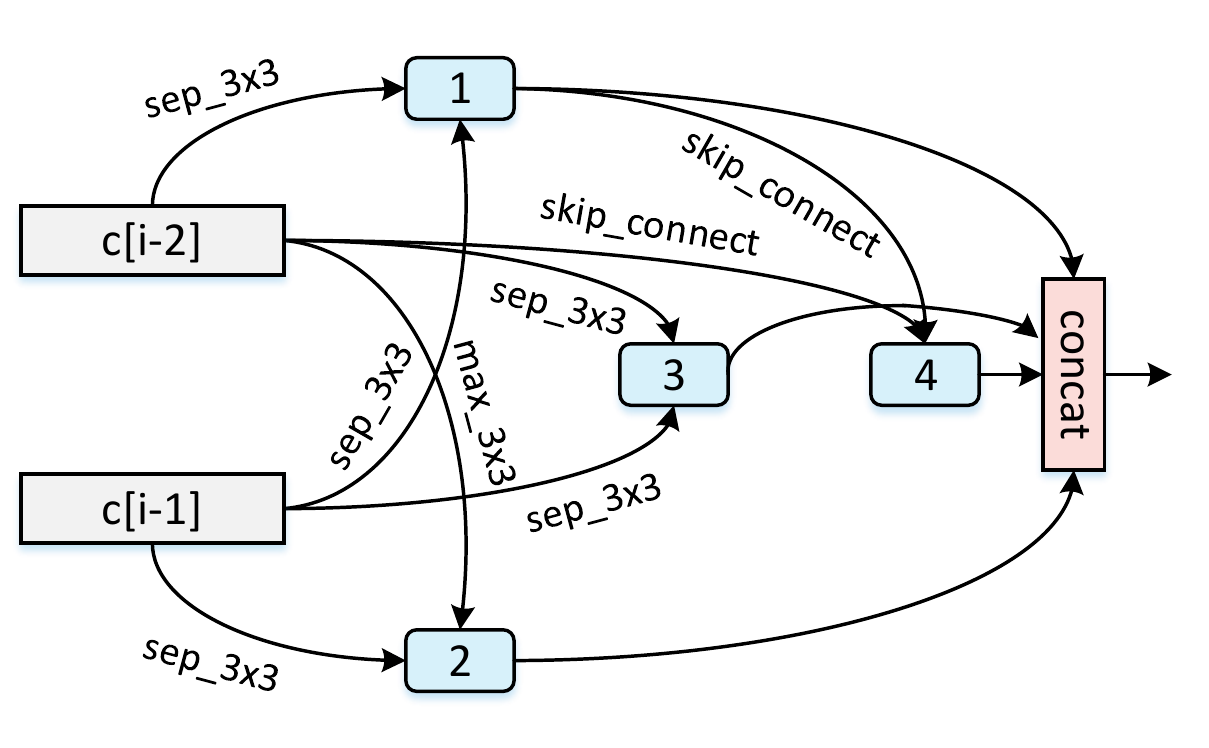}}
   \subfigure[Normal Cell by SDAS-B.]{
     \label{fig:cifar:b}
     \includegraphics[width=0.32\textwidth]{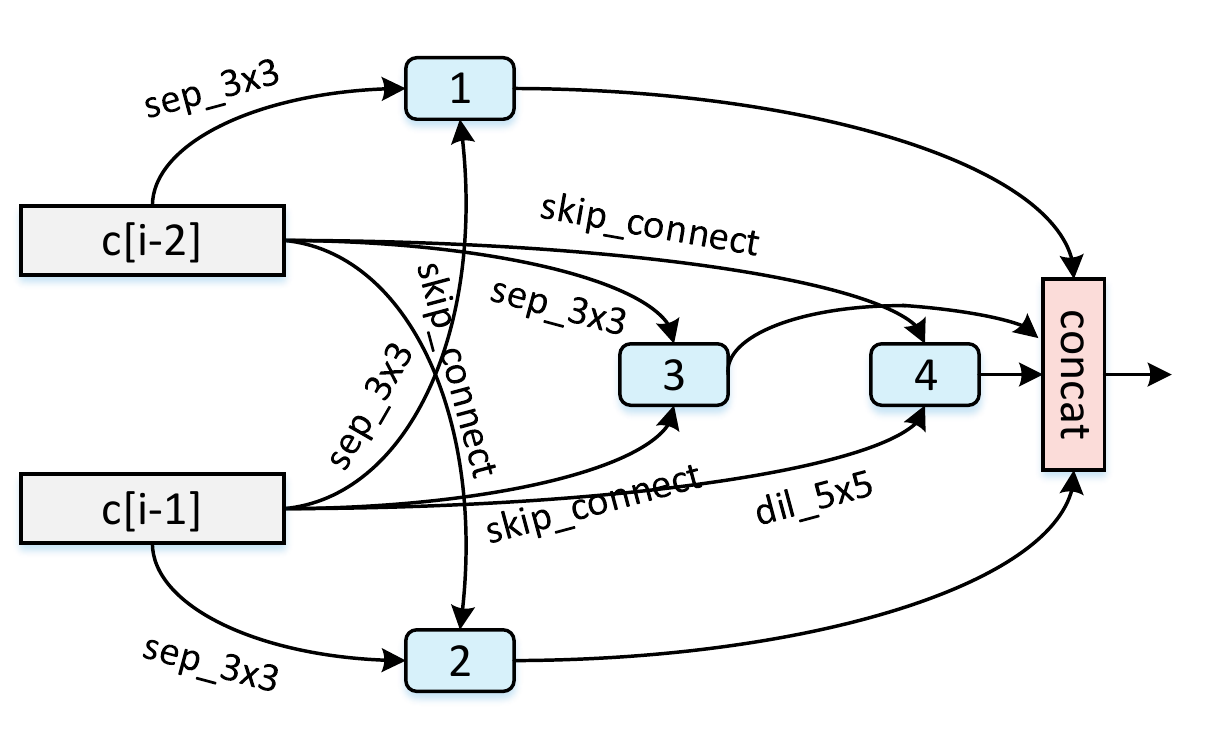}}
   \subfigure[Normal Cell by SDAS-C.]{
     \label{fig:cifar:c}
     \includegraphics[width=0.32\textwidth]{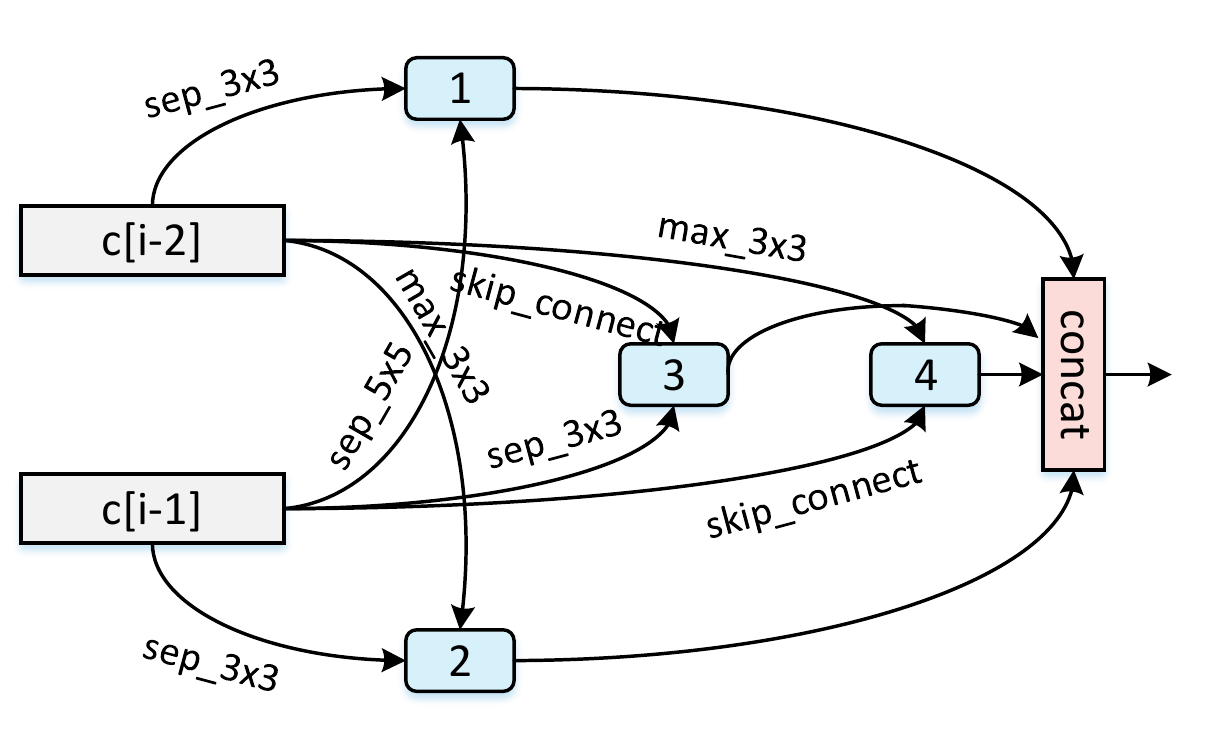}}
    \subfigure[Reduction Cell by SDAS-A.]{
     \label{fig:cifar:d}
     \includegraphics[width=0.32\textwidth]{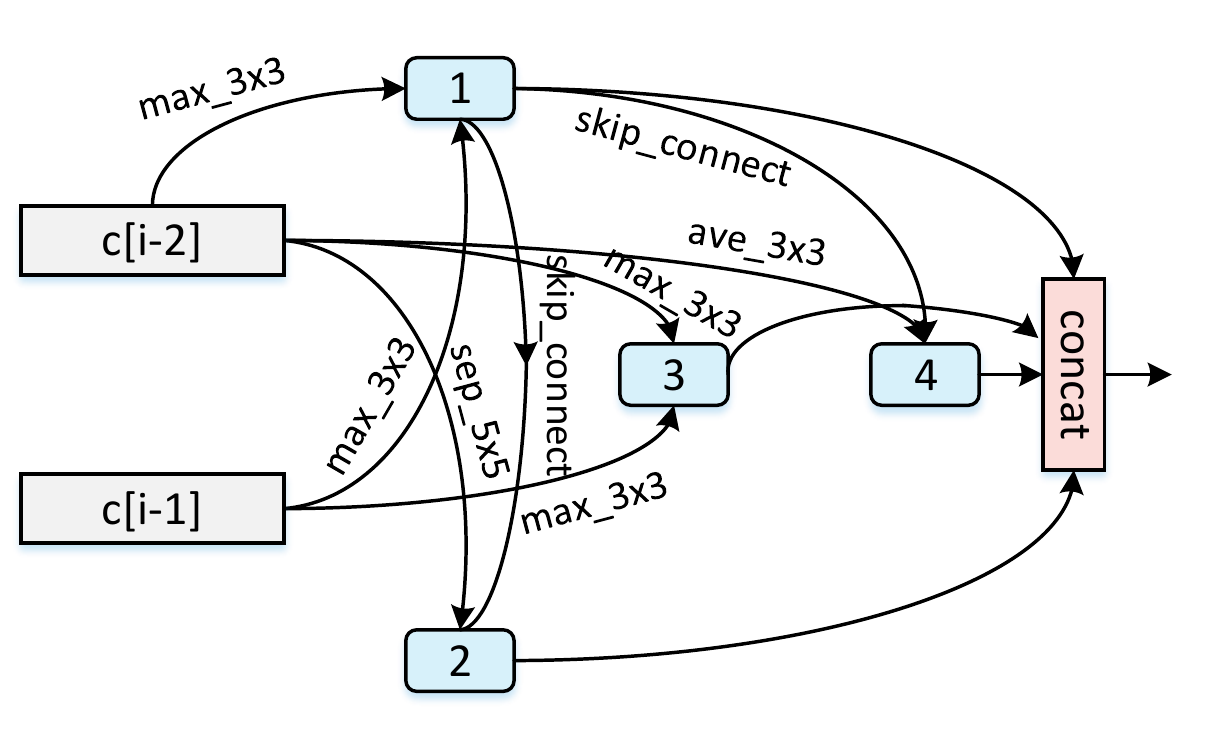}}
   \subfigure[Reduction Cell by SDAS-B.]{
     \label{fig:cifar:e}
     \includegraphics[width=0.32\textwidth]{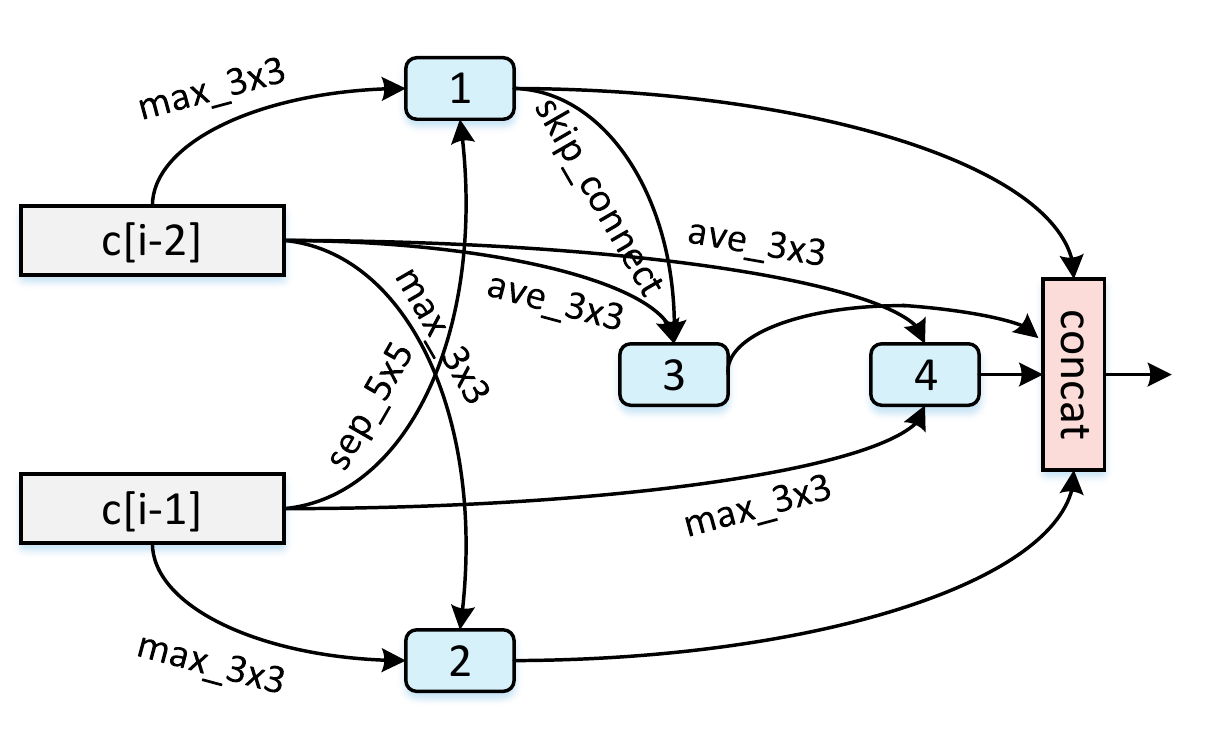}}
   \subfigure[Reduction Cell by SDAS-C.]{
     \label{fig:cifar:f}
     \includegraphics[width=0.32\textwidth]{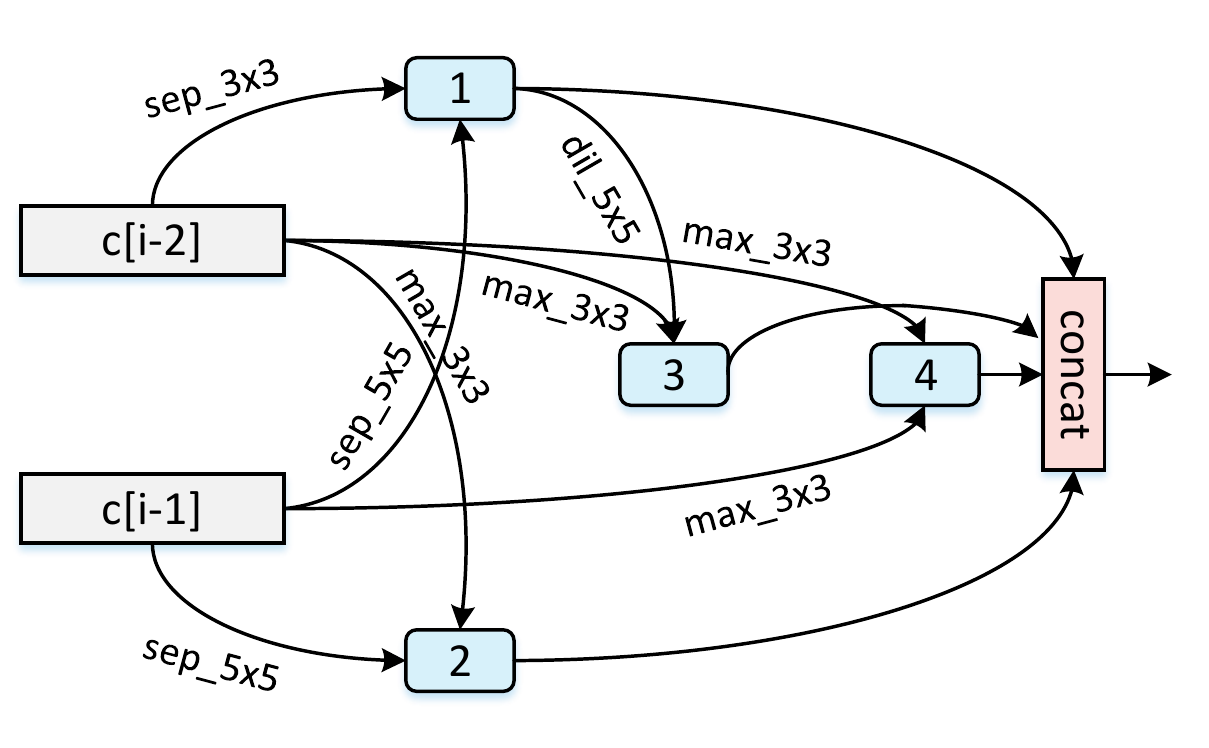}}
     \vspace{-0.10in}
   \caption{\small The optimal architectures or cells by SDAS with different schedule functions on CIFAR10. The input nodes (gray) are the output of two prior cells. The output (pink) is the result of a concatenation operation across all resulting branches. Each intermediate node corresponds to two operations and a combination operation (blue).}
   \label{fig:cifar}
   \vspace{-0.10in}
\end{figure*}

\begin{table*}[]
\centering
\small
\caption{Comparisons with state-of-the-art image classifiers on ImageNet in the mobile setting.}
\begin{tabular}{@{}lccccccc@{}}\hline
\multirow{2}{*}{\textbf{Architecture}}                & \multicolumn{2}{c}{\textbf{Test Error (\%)}} & \textbf{Params} & \textbf{$+ \times$} & \textbf{Search Cost} & \textbf{Search} \\ \cline{2-3}
                                                      & top-1 & top-5                                & \textbf{(M)}    & \textbf{(M)}        & \textbf{(GPU days)}  & \textbf{Method} \\ \hline
 Inception-v1 \citep{szegedy2015going}                 & 30.2  & 10.1                                 & 6.6             & 1448                & --                   & manual \\
 MobileNet \citep{howard2017mobilenets}                & 29.4  & 10.5                                 & 4.2             & 569                 & --                   & manual \\
 ShuffleNet 2$\times$ (v1) \citep{zhang2018shufflenet} & 29.1  & 10.2                                 & $\sim$5         & 524                 & --                   & manual \\
 ShuffleNet 2$\times$ (v2) \citep{zhang2018shufflenet} & 26.3  & --                                   & $\sim$5         & 524                 & --                   & manual \\
 NASNet-A \citep{zoph2018learning}                     & 26.0  & 8.4                                  & 5.3             & 564                 & 1800                 & RL \\
 NASNet-B \citep{zoph2018learning}                     & 27.2  & 8.7                                  & 5.3             & 488                 & 1800                 & RL \\
 NASNet-C \citep{zoph2018learning}                     & 27.5  & 9.0                                  & 4.9             & 558                 & 1800                 & RL \\
 AmoebaNet-A \citep{real2018regularized}               & 25.5  & 8.0                                  & 5.1             & 555                 & 3150                 & evolution \\
 AmoebaNet-B \citep{real2018regularized}               & 26.0  & 8.5                                  & 5.3             & 555                 & 3150                 & evolution \\
 AmoebaNet-C \citep{real2018regularized}               & 24.3  & 7.6                                  & 6.4             & 570                 & 3150                 & evolution \\
 PNAS \citep{liu2018progressive}                       & 25.8  & 8.1                                  & 5.1             & 588                 & $\sim$225            & SMBO \\ \hline
 DARTS                       \citep{liu2018darts}      & 26.9  & 9.0                                  & 4.9             & 595                 & 4                    & gradient-based \\
 SNAS + mild                 \citep{xie2018snas}       & 27.3  & 9.2                                  & 4.3             & 522                 & 1.5                  & gradient-based \\ \hline
 SDAS-A                                       & 27.0  & 9.1                                  & 4.6             & 531                  & 0.31                  & gradient-based  \\
 SDAS-B                                       & 26.6  & 8.7                                  & 4.4             & 510                  & 0.37                  & gradient-based  \\
 SDAS-C                                       & 26.7  & 8.8                                  & 4.5             & 511                  & 0.25                  & gradient-based  \\
 \hline
\end{tabular}
\vspace{-0.10in}
\label{tab:imagenet}
\end{table*}

Table \ref{tab:cifar10} summarizes the comparisons with state-of-the-art network architectures on CIFAR10 dataset. SDAS-A, SDAS-B and SDAS-C denote our SDAS with the three different schedule functions introduced in Section \ref{sec:sd}. For each DAS or SDAS setting, we run 4 times and report the average test error. Though DAS and SDAS both involve architecture discretization, they are different in the way that DAS is as a result of determining all the operations at one step, while SDAS is by selecting the operations during training with a schedule. As indicated by our results, utilizing a schedule in SDAS can constantly lead to a lower error rate than one-step decision in DAS. In particular, the SDAS-B achieves the lowest error across SDASs with different schedule functions, which demonstrates a possible $1.4\times$ speedup with 6.5\% relative drop in average test error over DAS. The optimal architectures learnt by SDASs with different schedule functions are shown in Figure \ref{fig:cifar}.

\begin{figure*}[!tb]
   \centering
   \subfigure[Normal Cell.]{
     \label{fig:k10:a}
     \includegraphics[width=0.32\textwidth]{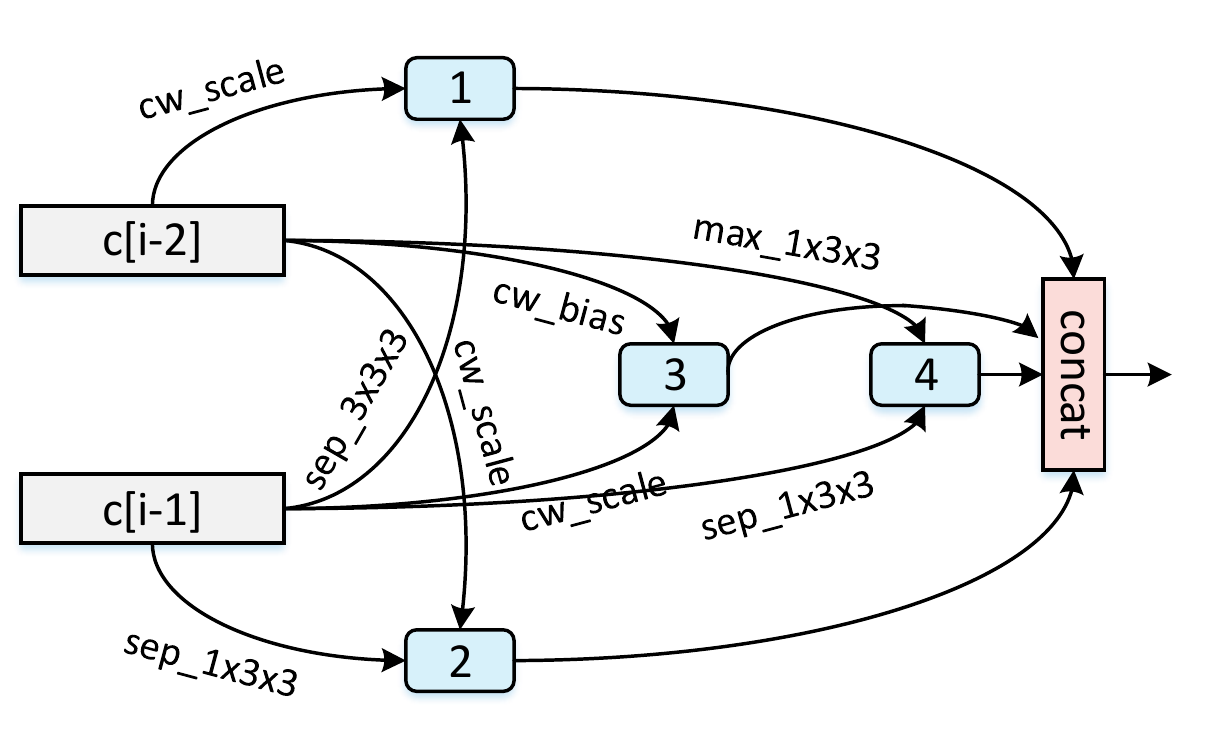}}
   \subfigure[ST-Reduction Cell.]{
     \label{fig:k10:b}
     \includegraphics[width=0.32\textwidth]{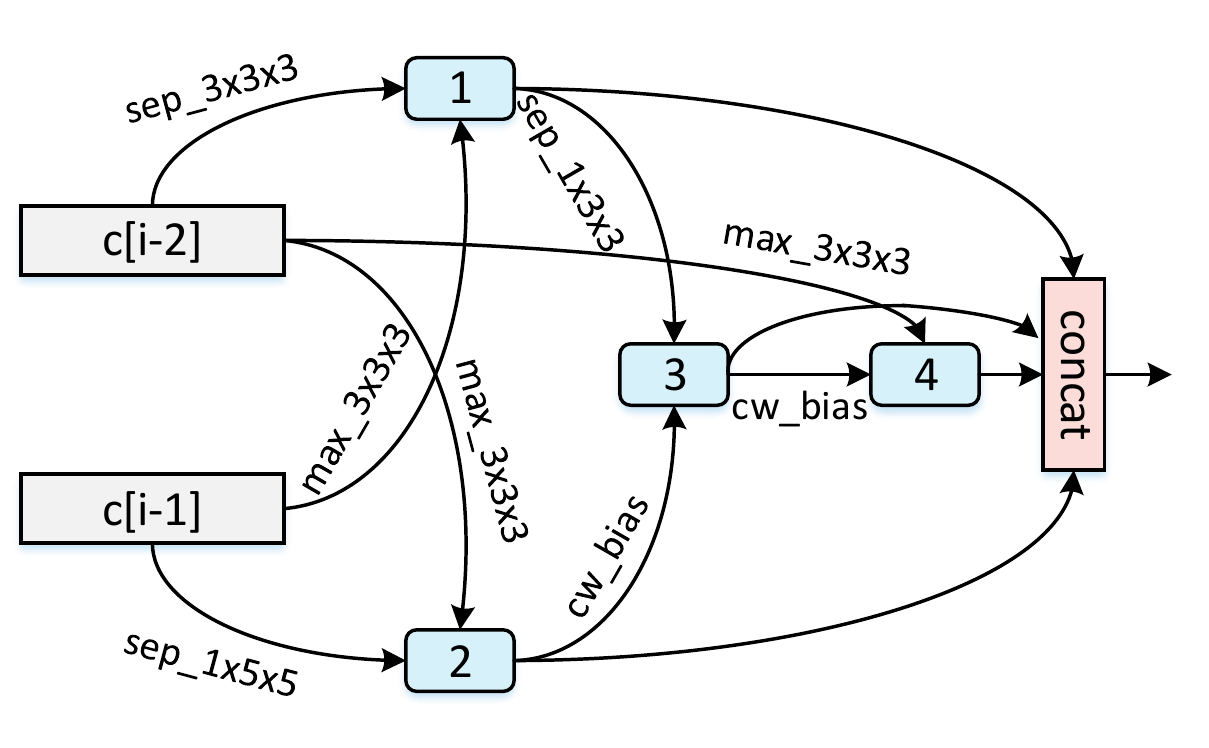}}
   \subfigure[Reduction Cell.]{
     \label{fig:k10:c}
     \includegraphics[width=0.32\textwidth]{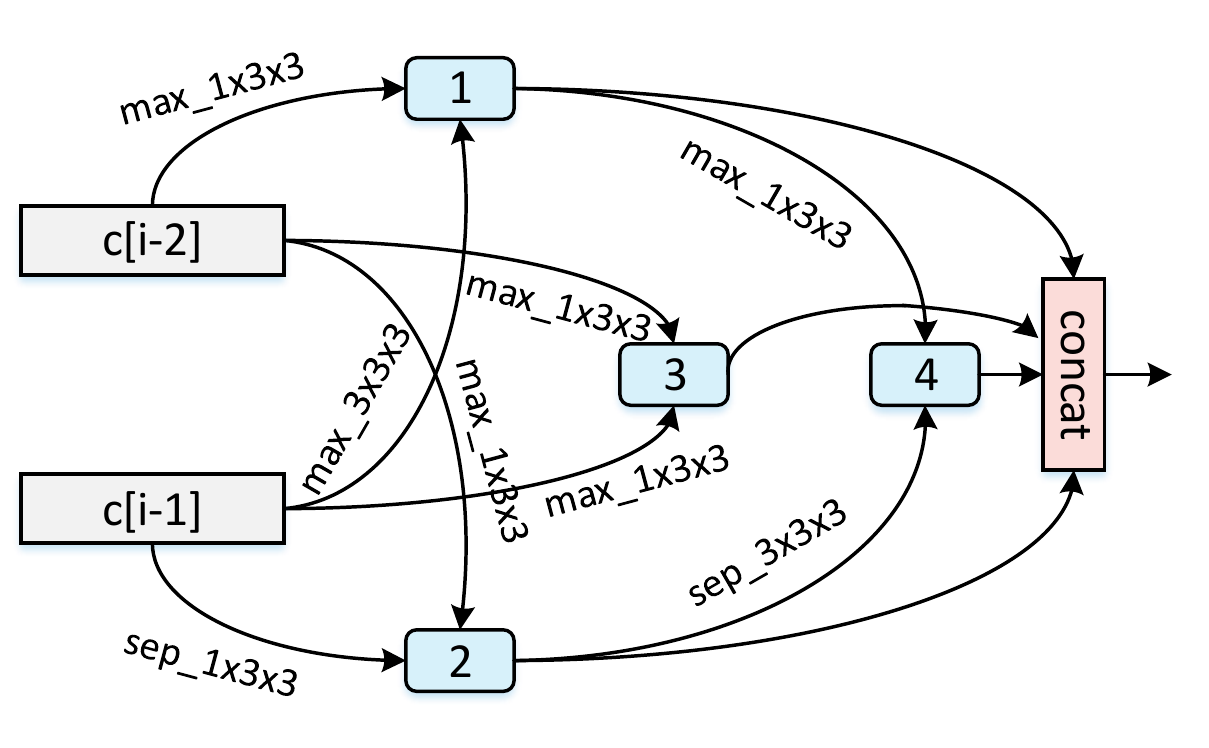}}
\vspace{-0.10in}
   \caption{\small The optimal architectures or cells by SDAS-C for video recognition on Kinetics10. The input nodes (gray) are the output of two prior cells. The output (pink) is the result of a concatenation operation across all resulting branches. Each intermediate node corresponds to two operations and a combination operation (blue).}
   \label{fig:k10}
   \vspace{-0.10in}
\end{figure*}

\subsection{Transferability of Learnt Architectures}

For the transferability of learnt architectures, following \citep{liu2018darts}, we apply the optimal architectures searched with each schedule to ImageNet dataset. Here, we only consider the mobile setting where the number of multiply-add operations with $224\times 224$ input is restricted to be less than 600M. In order to reach the close network scale of this limitation, the repeat number $K$ of Normal Cell is set as 4, and the output channels are fixed as $C_1=24$ and $C_2=192$ for the ImageNet evaluation. The network is trained for 250 epochs with batch size 128, weight decay $3\times 10^{-5}$ and initial learning rate 0.1 which is decayed by 0.97 after each epoch. The comparisons with state-of-the-art image classifiers on ImageNet in the mobile setting are summarized in Table \ref{tab:imagenet}. Similarly to the conclusion on CIFAR10, SDAS-B achieves $8.7\%$ top-5 test error, which is the lowest across three SDAS settings. When compared with DARTS and SNAS, which utilize the same search space as our setting, SDAS-B leads to $10\times$ and $4\times$ faster search speed with 3.3\% and 5.4\% relative drop on top-5 error, respectively. For the baselines NASNet, AmoebaNet and PNAS, which need much more computation resources during search, SDAS-B can also achieve comparable performance with even less parameters. This result basically indicates the transferability of learnt architectures by SDAS for image recognition.

\section{Experiments on Video Recognition}

\begin{table}
\centering
\small
\caption{\small Comparisons between performing search with different operation sets on Kinetics10.}
\vspace{0.1cm}
\begin{tabular}{lcccc} \hline
\multirow{2}*{\textbf{Method}} & \multirow{2}*{\textbf{Ops}} & \textbf{Search Cost} & \textbf{Params} & \textbf{Error} \\
                               &                             & \textbf{(GPU days)}           & \textbf{(M)} & \textbf{(\%)}  \\ \hline
\begin{minipage}{1.3cm}\vspace{0.1cm}DAS \vspace{0.1cm}\end{minipage} & $\mathcal{O}_{2D}$      & 6.8 & 1.41 & 11.67 $\pm$ 0.21 \\
\begin{minipage}{1.3cm}\vspace{0.1cm}DAS \vspace{0.1cm}\end{minipage} & $\mathcal{O}_{3D}$      & 9.9 & 1.39 & 10.88 $\pm$ 0.11 \\
\begin{minipage}{1.3cm}\vspace{0.1cm}DAS \vspace{0.1cm}\end{minipage} & $\mathcal{O}_{adv}$     & 9.2 & 1.46 & 11.49 $\pm$ 0.30 \\ \hline
\begin{minipage}{1.3cm}\vspace{0.1cm}SDAS-A \vspace{0.1cm}\end{minipage} & $\mathcal{O}_{2D}$   & 3.8 & 0.95 & 11.02 $\pm$ 0.08 \\
\begin{minipage}{1.3cm}\vspace{0.1cm}SDAS-A \vspace{0.1cm}\end{minipage} & $\mathcal{O}_{3D}$   & 5.7 & 1.06 & 10.68 $\pm$ 0.11 \\
\begin{minipage}{1.3cm}\vspace{0.1cm}SDAS-A \vspace{0.1cm}\end{minipage} & $\mathcal{O}_{adv}$  & 4.6 & 1.21 & 10.30 $\pm$ 0.19 \\ \hline
\begin{minipage}{1.3cm}\vspace{0.1cm}SDAS-B \vspace{0.1cm}\end{minipage} & $\mathcal{O}_{2D}$   & 5.2 & 0.94 & 11.09 $\pm$ 0.11 \\
\begin{minipage}{1.3cm}\vspace{0.1cm}SDAS-B \vspace{0.1cm}\end{minipage} & $\mathcal{O}_{3D}$   & 7.5 & 1.12 & 10.62 $\pm$ 0.15 \\
\begin{minipage}{1.3cm}\vspace{0.1cm}SDAS-B \vspace{0.1cm}\end{minipage} & $\mathcal{O}_{adv}$  & 5.9 & 1.10 & 10.24 $\pm$ 0.08 \\ \hline
\begin{minipage}{1.3cm}\vspace{0.1cm}SDAS-C \vspace{0.1cm}\end{minipage} & $\mathcal{O}_{2D}$   & 2.3 & 0.82 & 10.93 $\pm$ 0.27 \\
\begin{minipage}{1.3cm}\vspace{0.1cm}SDAS-C \vspace{0.1cm}\end{minipage} & $\mathcal{O}_{3D}$   & 3.5 & 0.92 & 10.58 $\pm$ 0.12 \\
\begin{minipage}{1.3cm}\vspace{0.1cm}SDAS-C \vspace{0.1cm}\end{minipage} & $\mathcal{O}_{adv}$  & 3.3 & 0.93 & 10.22 $\pm$ 0.10 \\ \hline
\end{tabular}
\label{tab:opt}
\vspace{-0.15in}
\end{table}

\subsection{Evaluations on Kinetics10}
\textbf{Evaluation on Operation Set.} For video recognition, we first examine how the performance of automatic network design is affected when capitalizing on different operation sets. Here, we search the optimal architecture on Kinetics10 by DAS and SDAS on the three operation set $\mathcal{O}_{2D}$, $\mathcal{O}_{3D}$ and $\mathcal{O}_{adv}$, respectively. During search, the repeat number of Normal Cell is set as $K=2$, and output channels are fixed as $C_1=16$ and $C_2=64$. After architecture search, we enlarge the learning capacity of network by using $K=2$, $C_1=32$ and $C_2=128$ unless otherwise stated. The optimal architecture is trained for 96 epochs with batch size 64 on Kinetics10 training set.

Table \ref{tab:opt} summarizes the search time, the number of parameters and error rate on Kinetics10 validation set by DAS and SDAS on the three operation sets. For each architecture, the mean and standard deviation across 4 independent runs are given. Overall, the results across three operation sets consistently indicate that the network with the architectures learnt by SDAS leads to a performance boost and search speed-up against that by DAS. Furthermore, searching the cells on $\mathcal{O}_{3D}$ operation set also exhibits better performance than architecture optimization on $\mathcal{O}_{2D}$ set by both DAS and SDAS. That basically verifies the merit of 3D operations to encode spatio-temporal context in videos. One observation is that DAS cannot handle the advanced operation well and achieves worse performance when switching to advanced set $\mathcal{O}_{adv}$. While for our SDAS, the involving of advanced operation can consistently boost up the performance, which demonstrate the ability of SDAS to deal with different kinds of operations.

Figure \ref{fig:k10} depicts the best performing Normal Cell, ST-Reduction Cell and S-Reduction Cell on Kinetics10 by SDAS-C and the advanced operation set is exploited in this experiments. An interesting observation is that the operation of 3d max pooling and 2d max pooling is often chosen in ST-Reduction Cell and S-Reduction Cell, respectively. This reasonably meets our expectation, which is in conformity with the manually designed networks.

\begin{table}
\centering
\small
\caption{\small The effect of free parameters $K$, $C_1$ and $C_2$ in network structure on Kinetics10.}
\vspace{0.1cm}
\begin{tabular}{l|c|ccc|c|c} \hline
\multirow{2}*{\textbf{Method}} & \multirow{2}*{\textbf{Ops}} & \multirow{2}*{$K$} & \multirow{2}*{$C_1$} & \multirow{2}*{$C_2$} & \textbf{Params}  & \textbf{Error} \\
                               &                             &                    &                      &                      & \textbf{(M)} & \textbf{(\%)}  \\ \hline
\multirow{5}*{SDAS-C} & \multirow{5}*{$\mathcal{O}_{adv}$} & 2 & 16 & 64  & 0.24 & 12.44 $\pm$ 0.19 \\
& & 2 & 24 & 96  & 0.53 & 11.11 $\pm$ 0.17 \\
& & 2 & 32 & 128  & 0.93 & 10.22 $\pm$ 0.10 \\
& & 3 & 32 & 128  & 1.31 & 10.08 $\pm$ 0.04 \\
& & 3 & 32 & 192  & 2.89 & 9.66 $\pm$ 0.06 \\
\hline
\end{tabular}
\label{tab:param}
\vspace{-0.15in}
\end{table}

\begin{figure*}[!tb]
   \centering {\includegraphics[width=0.94\textwidth]{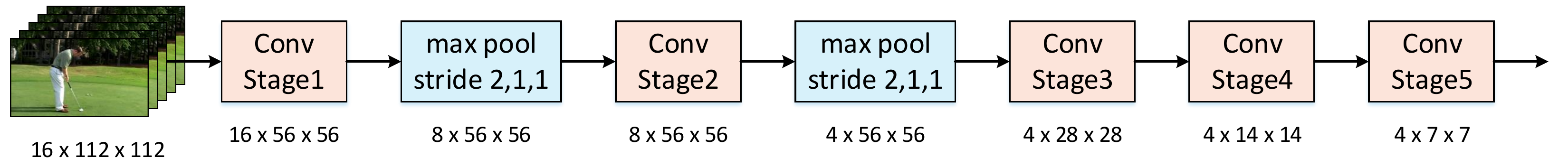}}
   \caption{\small The schematic illustration of extending hand-crafted structures to support the input of video clips. The network structure can be divided into five stages with respect to the change of resolution. One temporal max pooling is inserted after stage1 and stage2, respectively.}
   \label{fig:baselines}
   \vspace{-0.12in}
\end{figure*}

\textbf{The Effect of Free Parameters.} A common practice in architecture design that is tailored to network capacity is to consider the repeat number $K$ of Normal Cell and the output channels $C_1$ and $C_2$ in the stem layers and the first cell as free parameters. In the previous experiments, the values were optimally set as $K=2$, $C_1=32$ and $C_2=128$ to evaluate the network structure. Furthermore, we conducted the experiments to examine the effect of these parameters. Table \ref{tab:param} shows the comparisons between network structures by SDAS-C on Kinetics10 but exploiting different parameters. In general, the increase of $K$, $C_1$ and $C_2$ results in more parameters, but meanwhile the error rate decreases when using more cells and larger output channels. For example, the error drops in the range of 2.78\%.

\begin{table}
\centering
\small
\caption{\small Comparisons with hand-crafted structures on Kinetics10.}
\vspace{0.1cm}
\begin{tabular}{lccc} \hline
\multirow{2}*{\textbf{Method}} &   \textbf{Params} & $+\times$   & \textbf{Error} \\
                               &   \textbf{(M)}  & \textbf{(G)}& \textbf{(\%)}  \\ \hline
BNInception-2D  & 0.62 & 0.92 & 12.67 $\pm$ 0.20 \\
Xception-2D     & 1.28 & 1.15 & 11.31 $\pm$ 0.04 \\
Res50-2D        & 1.41 & 1.13 & 12.07 $\pm$ 0.05 \\
Res101-2D       & 2.55 & 2.00 & 11.78 $\pm$ 0.22 \\
ResNeXt50-2D    & 1.63 & 1.29 & 11.30 $\pm$ 0.18 \\
ResNeXt101-2D   & 3.01 & 2.34 & 11.30 $\pm$ 0.20 \\
SENet-2D        & 3.30 & 2.34 & 10.84 $\pm$ 0.10 \\ \hline
BNInception-3D  & 0.76 & 1.00 & 12.32 $\pm$ 0.15 \\
Xception-3D     & 1.30 & 1.17 & 10.61 $\pm$ 0.13 \\
Res50-3D        & 1.63 & 1.30 & 11.74 $\pm$ 0.27 \\
Res101-3D       & 2.97 & 2.32 & 11.33 $\pm$ 0.10 \\
ResNeXt50-3D    & 1.74 & 1.38 & 10.96 $\pm$ 0.21 \\
ResNeXt101-3D   & 3.22 & 2.50 & 10.40 $\pm$ 0.05 \\
SENet-3D        & 3.51 & 2.50 & 10.03 $\pm$ 0.09 \\ \hline
SDAS-C$_{2\_16\_64}$   & 0.24 & 0.36 & 12.44 $\pm$ 0.19 \\
SDAS-C$_{2\_24\_96}$   & 0.53 & 0.79 & 11.11 $\pm$ 0.17 \\
SDAS-C$_{2\_32\_128}$   & 0.93 & 1.38 & 10.22 $\pm$ 0.10 \\
SDAS-C$_{3\_32\_128}$   & 1.31 & 1.52 & 10.08 $\pm$ 0.04 \\
SDAS-C$_{3\_32\_192}$   & 2.89 & 2.20 & 9.66 $\pm$ 0.06 \\
\hline
\end{tabular}
\label{tab:hand-crafted}
\vspace{-0.15in}
\end{table}

\begin{figure}[!tb]
   \centering
   \subfigure[Accuracy versus number of parameters.]{
     \label{fig:curve:a}
     \includegraphics[width=0.4\textwidth]{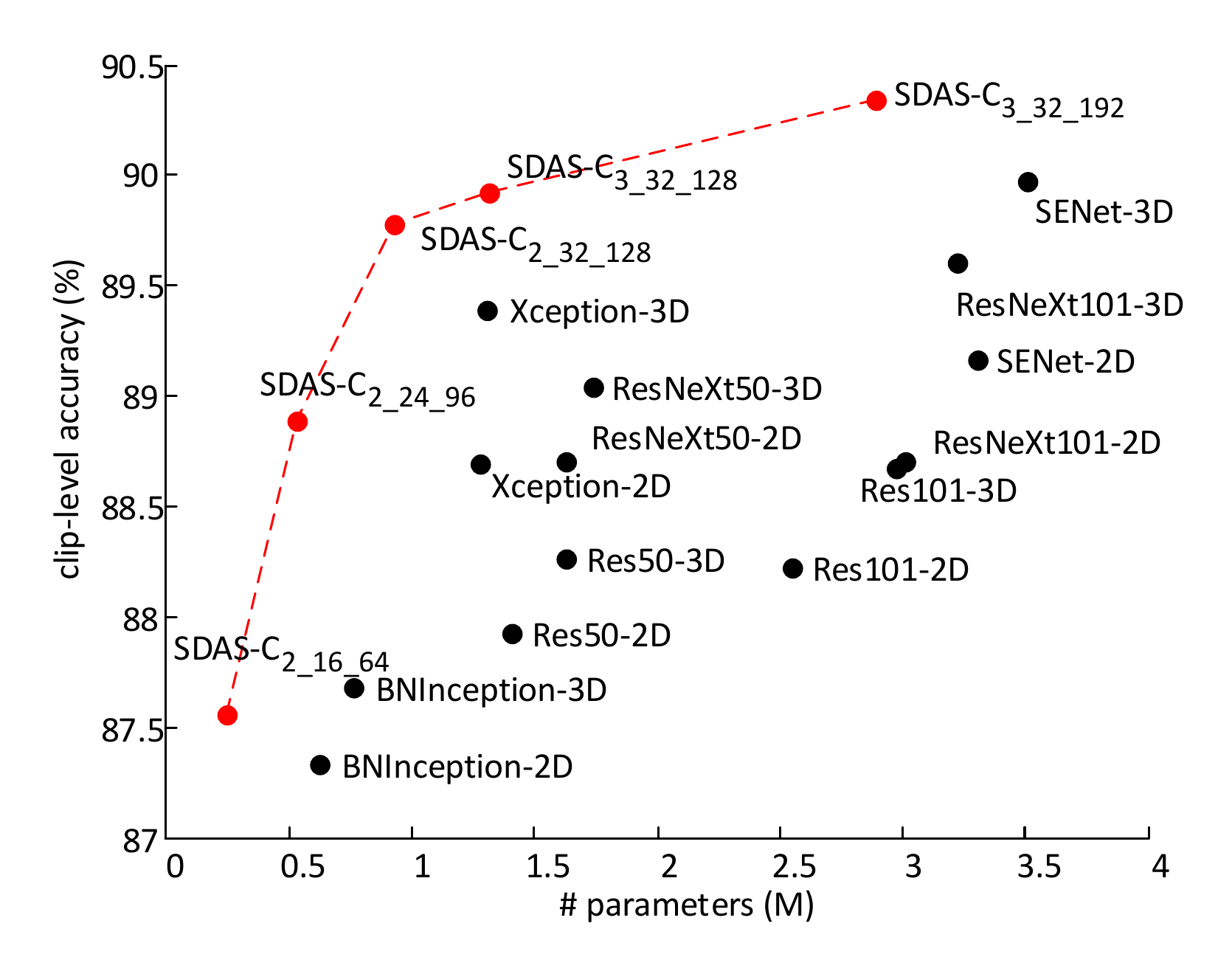}}
   \subfigure[Accuracy versus number of multiply-add operations.]{
     \label{fig:curve:b}
     \includegraphics[width=0.4\textwidth]{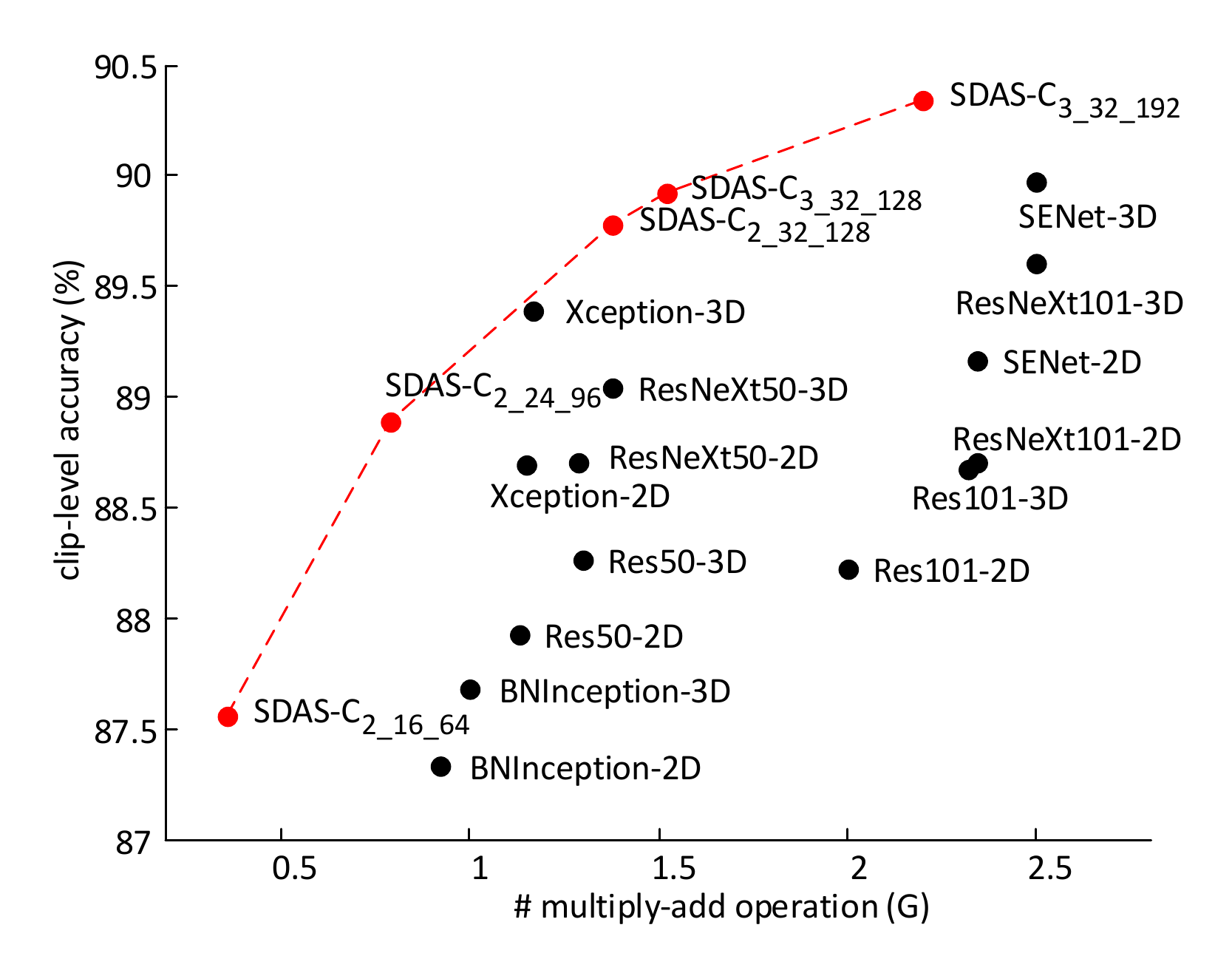}}
   \caption{\small Comparisons with hand-crafted structures on Kinetics10. Black/red circles indicate hand-crafted/auto-designed networks. SDAS-C$_{K, C_1, C_2}$ denotes the network structure with free parameters $K$, $C_1$ and $C_2$.}
   \label{fig:curve}
   \vspace{-0.10in}
\end{figure}

\textbf{Comparison with Hand-Crafted Structures.} We compare the following hand-crafted network structures for architecture evaluation: BNInception \citep{ioffe2015batch}, ResNet \citep{he2015deep}, ResNeXt \citep{xie2017aggregated}, Xception \citep{Chollet2017CVPR} and SENet \citep{hu2018squeeze}. To support the short-clip input in these 2D CNN structures, as shown in Figure \ref{fig:baselines}, we change 2D convolutions to $1\times k_s \times k_s$ spatial convolution and insert two temporal max pooling to reduce the temporal dimension progressively. These structures can also be extended to 3D networks by attaching a parallel $3\times1\times1$ temporal convolution to each spatial convolution. To make the scale of each structure comparable with our SDAS on Kinetics10, we reduce the output channels of layers in these networks by a factor of 4.

Table \ref{tab:hand-crafted} compares the architectures by SDAS-C with the hand-crafted network structures. The results are also shown in Figure \ref{fig:curve} in the form of accuracy-vs-\#param curve and accuracy-vs-\#operation curve for better view. The network structures on the cells learnt by our SDAS-C across different scales consistently achieve superior accuracy to that of human-invented architectures with comparable number of parameters or comparable number of multiply-add operations. Though both our $\mathcal{O}_{adv}$ set and SENet involve utilization of the channels-wise operation, the network structure by SDAS-C is benefited from the way of automatic design and leads to better result.

\begin{figure*}[!tb]
   \centering
   \subfigure[Normal Cell.]{
     \label{fig:ucf101:a}
     \includegraphics[width=0.32\textwidth]{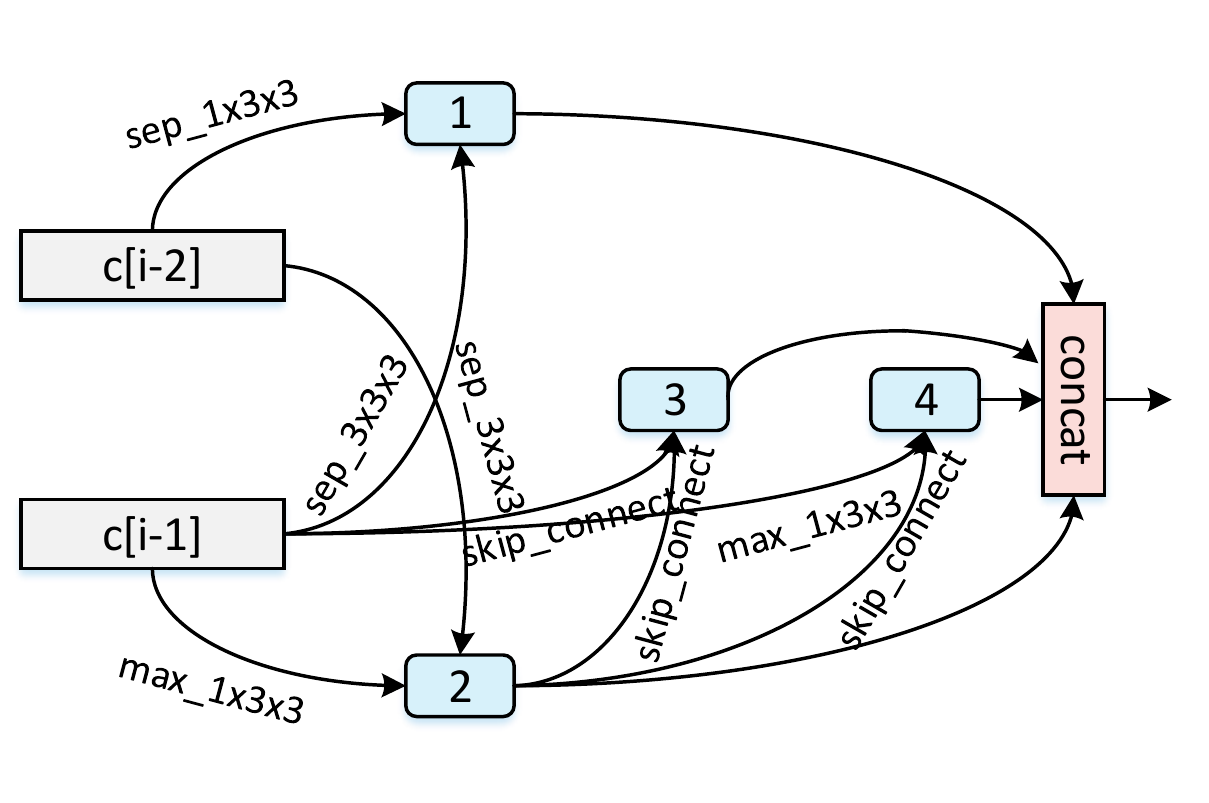}}
   \subfigure[ST-Reduction Cell.]{
     \label{fig:ucf101:b}
     \includegraphics[width=0.32\textwidth]{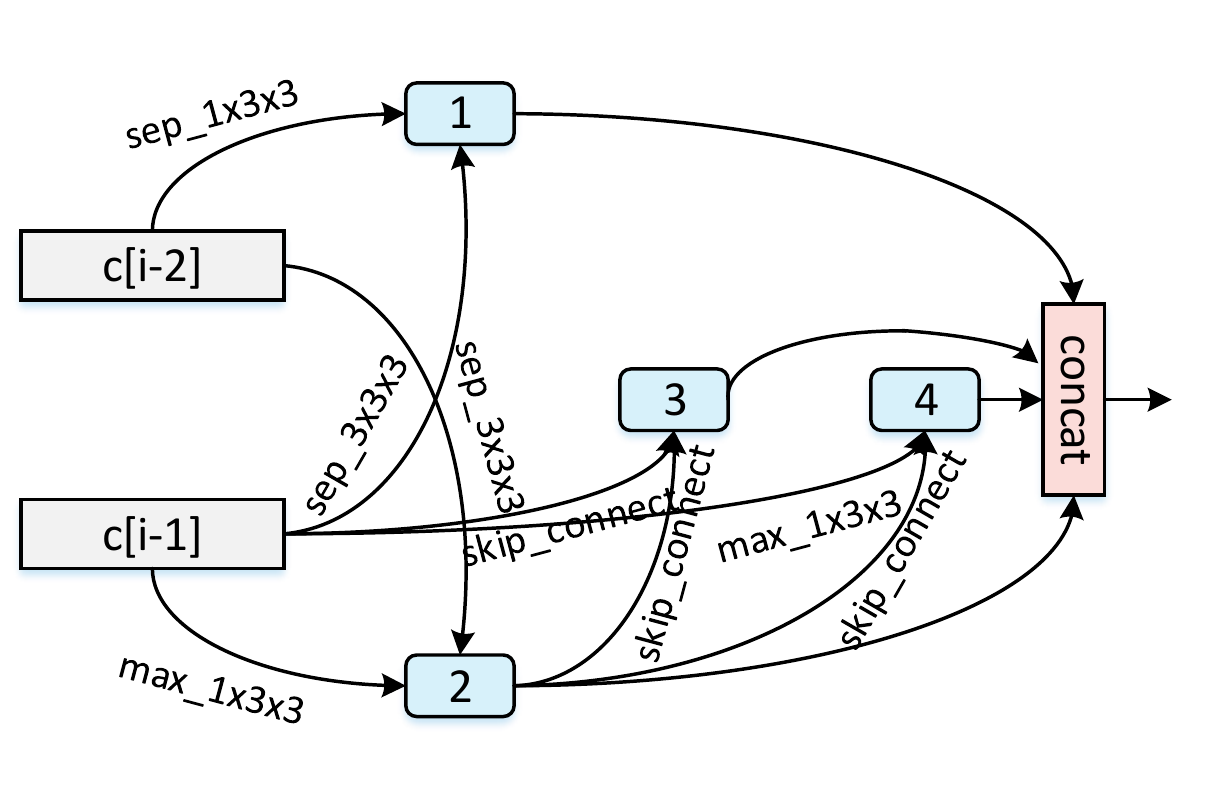}}
   \subfigure[Reduction Cell.]{
     \label{fig:ucf101:c}
     \includegraphics[width=0.32\textwidth]{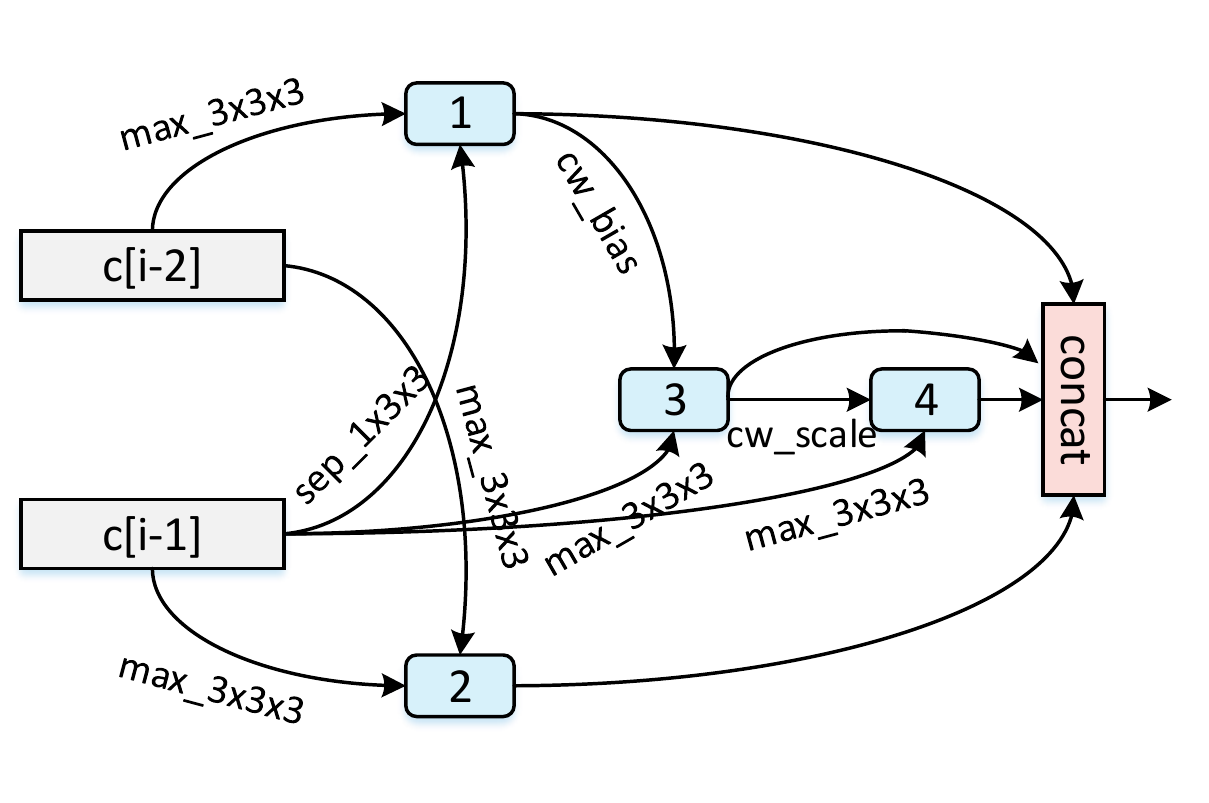}}
\vspace{-0.10in}
   \caption{\small The optimal architectures or cells by SDAS-C for video recognition on UCF101. The input nodes (gray) are the output of two prior cells. The output (pink) is the result of a concatenation operation across all resulting branches. Each intermediate node corresponds to two operations and a combination operation (blue).}
   \label{fig:ucf101}
\vspace{-0.10in}
\end{figure*}

\begin{figure*}[!tb]
   \centering
   \subfigure[Normal Cell.]{
     \label{fig:hmdb51:a}
     \includegraphics[width=0.32\textwidth]{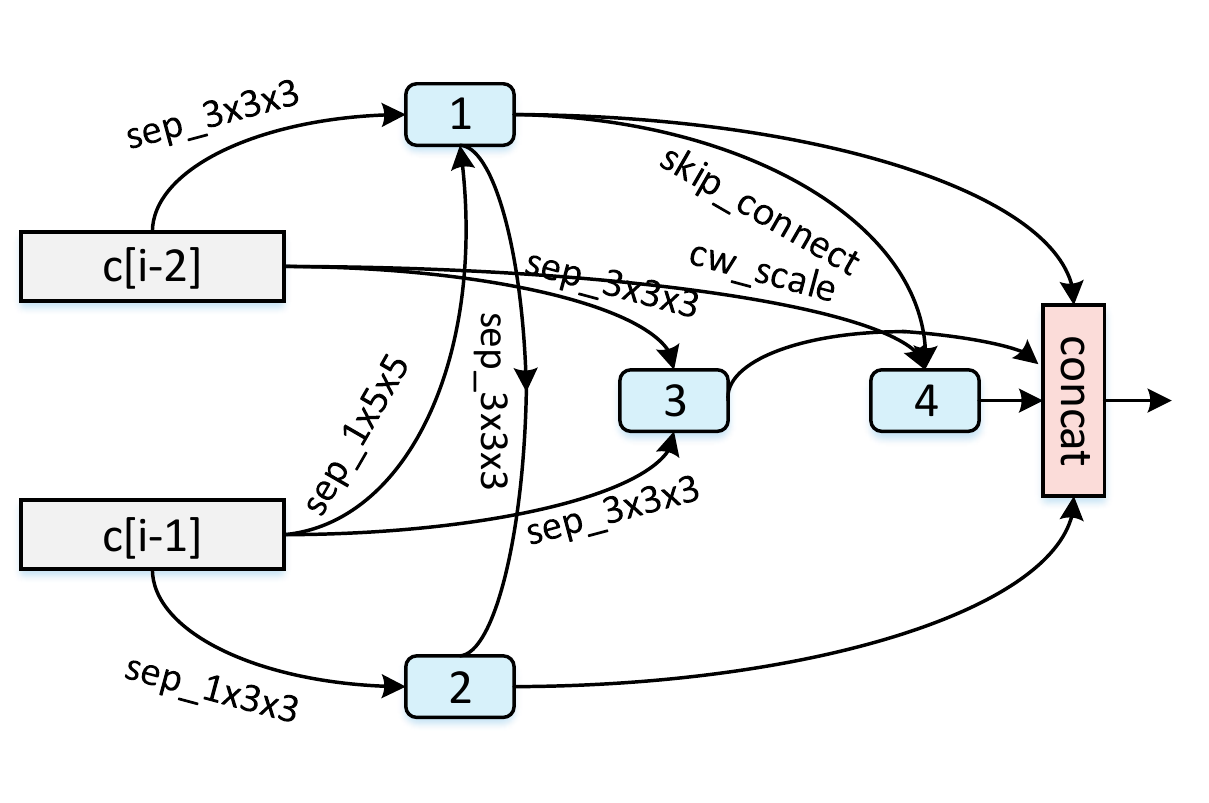}}
   \subfigure[ST-Reduction Cell.]{
     \label{fig:hmdb51:b}
     \includegraphics[width=0.32\textwidth]{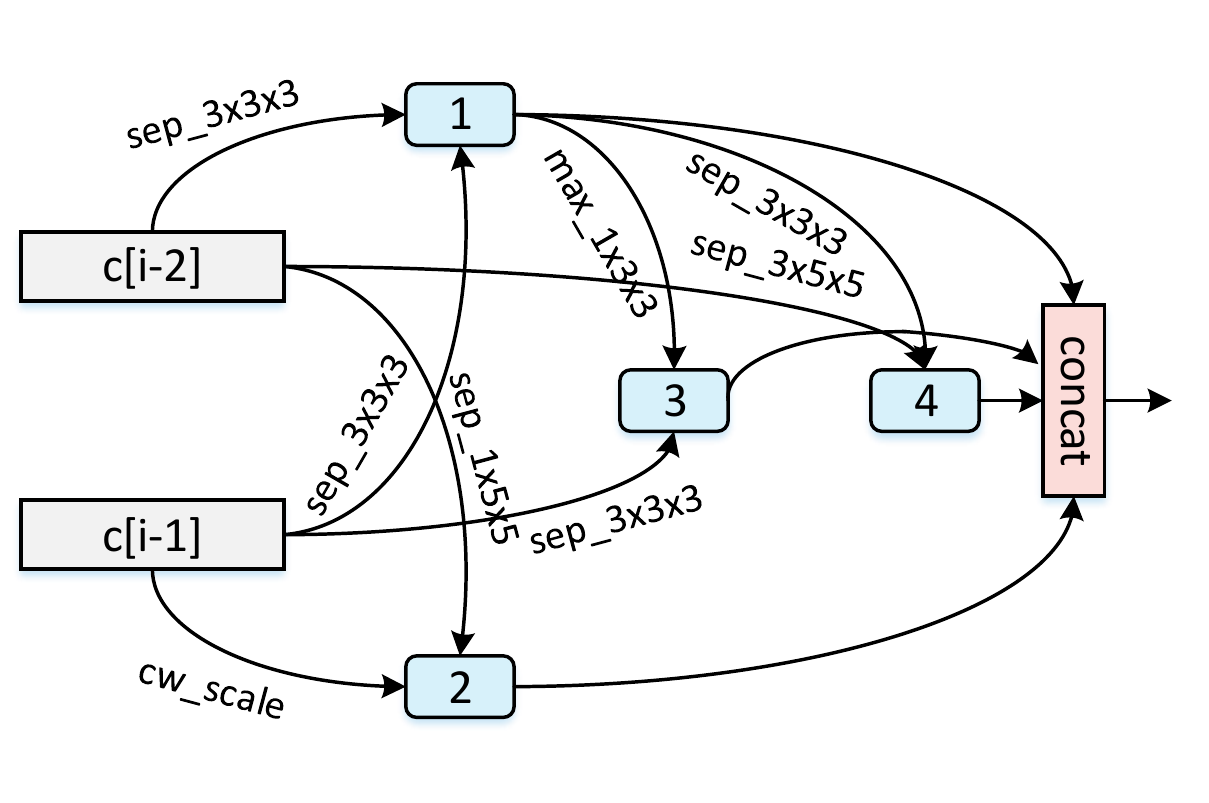}}
   \subfigure[Reduction Cell.]{
     \label{fig:hmdb51:c}
     \includegraphics[width=0.32\textwidth]{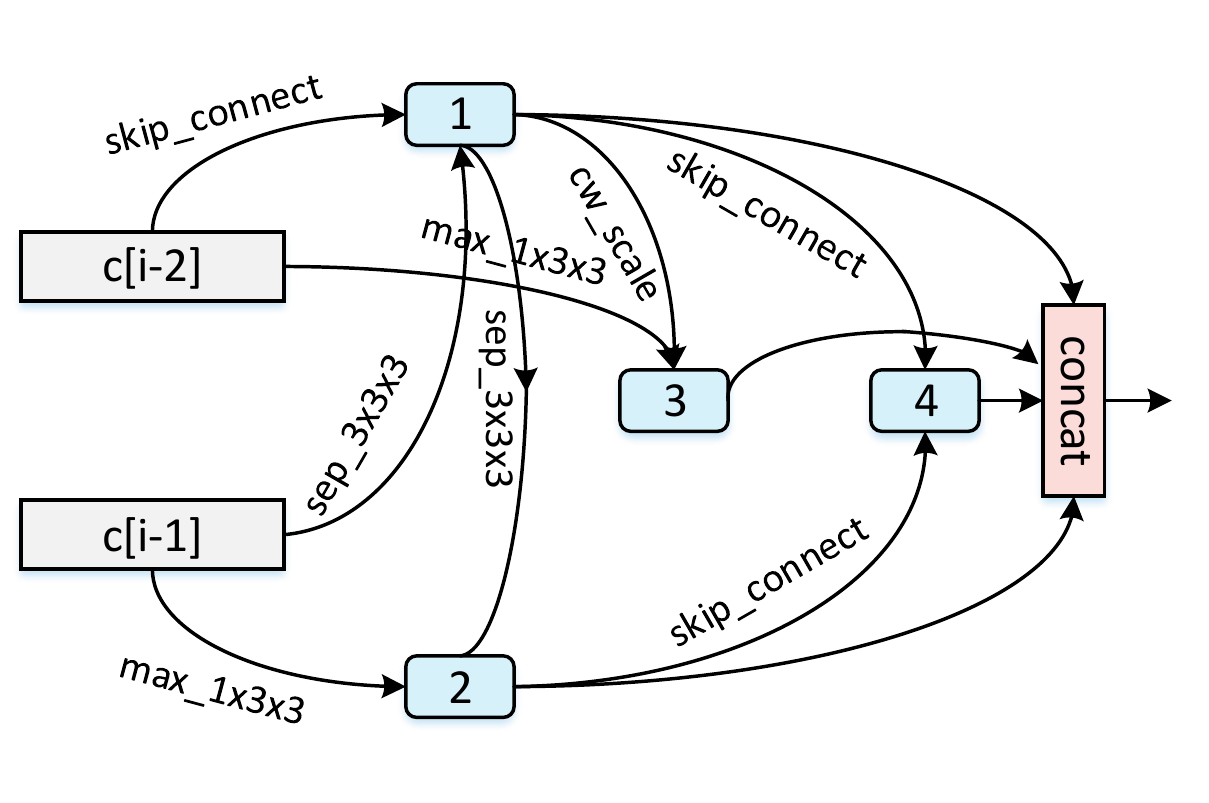}}
\vspace{-0.10in}
   \caption{\small The optimal architectures or cells by SDAS-C for video recognition on HMDB51. The input nodes (gray) are the output of two prior cells. The output (pink) is the result of a concatenation operation across all resulting branches. Each intermediate node corresponds to two operations and a combination operation (blue).}
   \label{fig:hmdb51}
\vspace{-0.10in}
\end{figure*}

\begin{table}
\centering
\small
\caption{\small Comparisons between DAS and SDAS-C when performing search on UCF101 and HMDB51.}
\vspace{0.1cm}
\begin{tabular}{lcccc} \hline
\multirow{2}*{\textbf{Method}} & \multirow{2}*{\textbf{Ops}} & \textbf{Search Cost} & \textbf{Params} & \textbf{Error} \\
                               &                             & \textbf{(GPU days)}           & \textbf{(M)} & \textbf{(\%)}  \\ \hline
\textbf{UCF101} \\ \hline
\begin{minipage}{1.3cm}\vspace{0.1cm}DAS \vspace{0.1cm}\end{minipage} & $\mathcal{O}_{adv}$      & 7.3 & 0.73 & 37.49 $\pm$ 0.91 \\
\begin{minipage}{1.3cm}\vspace{0.1cm}SDAS-C \vspace{0.1cm}\end{minipage} & $\mathcal{O}_{adv}$   & 2.6 & 0.77 & 37.19 $\pm$ 0.92 \\ \hline
\textbf{HMDB51} \\ \hline
\begin{minipage}{1.3cm}\vspace{0.1cm}DAS \vspace{0.1cm}\end{minipage} & $\mathcal{O}_{adv}$   & 2.7 & 0.91 & 74.08 $\pm$ 1.18 \\
\begin{minipage}{1.3cm}\vspace{0.1cm}SDAS-C \vspace{0.1cm}\end{minipage} & $\mathcal{O}_{adv}$  & 0.9 & 1.15 & 73.87 $\pm$ 0.25 \\ \hline

\end{tabular}
\label{tab:ucf101_hmdb51}
\vspace{-0.15in}
\end{table}

\subsection{Evaluations on UCF101 and HMDB51}
In this section, we turn to examine how the proposed SDAS behave when searching the optimal architecture on different datasets. Here, we utilize two famous action recognition datasets, i.e., UCF101 and HMDB51. For these two datasets, all the settings are the same as that for Kinetics10 except the number of iterations. Table \ref{tab:ucf101_hmdb51} compares the architectures by DAS and SDAS-C on UCF101 and HMDB51. The results on both datasets indicate that the network with the cells learnt by SDAS-C outperforms against that by DAS with $2.8\sim3\times$ search speedup. However, it should also be noticed that the standard deviation for each architecture on these two datasets is quite large ($0.25\sim 1.18$), which is close to the difference between architectures sometimes. As such, the experimental results are somewhat unstable when searching on UCF101 and HMDB51. Figure \ref{fig:ucf101} and Figure \ref{fig:hmdb51} shows the optimal architecture by SDAS-C on UCF101 and HMDB51, respectively.

\begin{table}
\centering
\small
\caption{\small Comparisons of transferability on Kinetics400 dataset.}
\vspace{0.1cm}
\begin{tabular}{l|c|c|c|c} \hline
\multirow{2}*{\textbf{Method}} & \textbf{Search} & \textbf{Search Cost} & \textbf{Params}  & \textbf{Top-1 Acc} \\
&                 \textbf{Dataset}                        &  \textbf{(GPU days)}    &        \textbf{(M)} & \textbf{(\%)}  \\ \hline
\multirow{3}*{DAS} & Kinetics10 & 9.2 & 37.68 & 73.3 \\
& UCF101 &  7.3 & 19.71 & 71.2 \\
& HMDB51 &  2.7 & 24.92  & 72.3 \\ \hline
\multirow{3}*{SDAS-C} & Kinetics10 & 3.3 & 26.01 & 74.2 \\
& UCF101 &  2.6 & 20.82  & 71.5\\
& HMDB51 &  0.9 & 31.07  & 73.0\\ \hline
\end{tabular}
\label{tab:cross}
\vspace{-0.15in}
\end{table}

\subsection{Transferability of Learnt Architectures}
To validate the transferability of learnt architectures, we perform a series of experiments on Kinetics400 with the best architectures searched from Kinetics10, UCF101 and HMDB51. Note that we merely transfer the architectures but train the weights of all models on Kinetics400 from scratch. The free parameters are set as $K=4$, $C_1=64$ and $C_2=512$. The network is trained for 128 epochs with batch size 256. Table \ref{tab:cross} summarizes the top-1 accuracy of different architectures on Kinetics400 validation set. As expected, SDAS-C using the cells learnt on Kinetics10 (a subset of Kinetics400) outperforms architectures searched on other dataset, or searched by DAS. For example, the architecture searched by SDAS-C on Kinetics10 leads to 1.2\% relative improvement against that by DAS with $45\%$ more parameters.

Then, we further extend the best performing architecture by SDAS-C with more than 16 input frames to model long-term temporal information, as summarized in Table \ref{tab:k400}. For the clip longer than 16 frames, we firstly train the networks with 16-frame clips and then fine-tune on target length of frames to speed up the optimization. In addition to RGB image input, we further execute the architecture search with optical flow image input to model the change of consecutive frames. Specifically, the network can capitalize on the two-direction optical flow images extracted by TV-L1 algorithm \citep{zach2007duality} by changing the input channels of first convolution to 2. As indicated by our results, the increase of the number of frames/optical flow images generally leads to performance improvements. On the input of RGB frames/optical flow images, the top-1 accuracy is boosted up from 74.2\%/64.8\% to 76.5\%/69.4\% when the number changes from 16 to 128. The accuracy of late fusion on two streams reaches 78.7\%.

\begin{table}
\centering
\small
\caption{\small Network evaluation on Kinetics400 dataset.}
\vspace{0.1cm}
\begin{tabular}{l|c|c|c|c} \hline
\multirow{2}*{\textbf{Method}} & \multirow{2}*{\textbf{Input}}& \multirow{2}*{\textbf{Frames}} & \textbf{Params}  & \textbf{Top-1 Acc} \\
                               &                             &                    &               \textbf{(M)} & \textbf{(\%)}  \\ \hline
\multirow{4}*{SDAS-C} & \multirow{4}*{RGB} & 16 & \multirow{4}*{26.01} & 74.2 \\
& & 32 &  & 75.1 \\
 & & 64 & & 76.0 \\
 & & 128 &  & 76.5 \\ \hline
\multirow{4}*{SDAS-C} & \multirow{4}*{Flow} & 16 & \multirow{4}*{26.01} & 64.8 \\
& & 32 &  & 67.4 \\
 & & 64 &  & 68.6 \\
 & & 128 &  & 69.4 \\ \hline
\multicolumn{2}{c|}{\begin{minipage}{2.8cm}\vspace{0.12cm}SDAS-C Two-stream \vspace{0.12cm}\end{minipage}} & 128 & 52.02 & 78.7 \\ \hline
\end{tabular}
\label{tab:k400}
\vspace{-0.15in}
\end{table}

\subsection{Performance Comparisons on Kinetics400}

We compare with the following state-of-the-art hand-crafted networks on Kinetics400 dataset and all the baselines are either trained on Kinetics400 from scratch or pre-trained on ImageNet or Sports1M dataset.

(1) Inflated 3D ConvNet (I3D) \citep{carreira2017quo} expands the 2D convolutions and 2D poolings in Inception network to 3D.

(2) Pseudo-3D ResNet (P3D) \citep{qiu2017learning} proposes a family of P3D blocks to decompose 3D learning into 2D convolutions in spatial space and 1D operations in temporal dimension.

(3) ResNet-200, ResNeXt-101, DenseNet-201 \citep{hara2018can} constructs a series of 3D networks by replacing the 2D filters with 3D filters in different CNN architectures. Here, we list there architectures deriving from ResNet \citep{he2015deep}, ResNeXt \citep{xie2017aggregated} and DenseNet \citep{huang2017densely}, respectively.

(4) R(2+1)D \citep{tran2018closer} builds spatio-temporal residual network with 2D spatial convolutions and 1D temporal convolutions. The R(2+1)D model is either trained from scratch or pre-trained on large-scale Sports1M \citep{karpathy2014large}~dataset.

(5) S3D-G \citep{xie2018rethinking} redesigns the Inception block used in I3D by replacing the inflated 3D kernel with one 2D convolution and one 1D convolution.

(6) NL I3D \citep{wang2018non} inserts the non-local operation between the residual blocks of ResNet101 to leverage the relation across different positions in the feature map.

\begin{table}
\centering
\small
\caption{\small Performance comparisons with the state-of-the-art methods on Kinetics400 validation set. }
\vspace{0.1cm}
\begin{tabular}{lccc} \hline
\multirow{2}*{\textbf{Method}} & \multirow{2}*{\textbf{Pre-training}} & \textbf{Top-1} & \textbf{Top-5} \\
 & & \textbf{(\%)} & \textbf{(\%)}\\ \hline
\begin{minipage}{3.0cm}\vspace{0.08cm}I3D \vspace{0.08cm}\end{minipage}& none & 67.5 & 87.2 \\
\begin{minipage}{3.0cm}\vspace{0.08cm}ResNet-200 \vspace{0.08cm}\end{minipage}& none & 63.1 & 84.4 \\
\begin{minipage}{3.0cm}\vspace{0.08cm}ResNeXt-101 \vspace{0.08cm}\end{minipage}& none & 65.1 & 85.7 \\
\begin{minipage}{3.0cm}\vspace{0.08cm}DenseNet-201  \vspace{0.08cm}\end{minipage}& none & 61.3 & 83.3 \\
\begin{minipage}{3.0cm}\vspace{0.08cm}R(2+1)D \vspace{0.08cm}\end{minipage}& none & 72.0 & 90.0 \\\hline
\begin{minipage}{3.0cm}\vspace{0.08cm}\textbf{SDAS-C} \vspace{0.08cm}\end{minipage}& none & 76.5 & 93.1 \\
\begin{minipage}{3.0cm}\vspace{0.08cm}\textbf{SDAS-C Two-stream} \vspace{0.08cm}\end{minipage}& none & \textbf{78.7} & \textbf{94.2} \\ \hline \hline

\begin{minipage}{3.0cm}\vspace{0.08cm}I3D \vspace{0.08cm}\end{minipage}& ImageNet & 72.1 & 90.3 \\
\begin{minipage}{3.0cm}\vspace{0.08cm}I3D Two-stream \vspace{0.08cm}\end{minipage}& ImageNet & 75.7 & 92.0 \\
\begin{minipage}{3.0cm}\vspace{0.08cm}P3D \vspace{0.08cm}\end{minipage}& ImageNet & 72.6 & 90.7 \\
\begin{minipage}{3.0cm}\vspace{0.08cm}R(2+1)D \vspace{0.08cm}\end{minipage}& Sports1M & 74.3& 91.4 \\
\begin{minipage}{3.0cm}\vspace{0.08cm}R(2+1)D Two-stream \vspace{0.08cm}\end{minipage}& Sports1M & 75.4& 91.9 \\
\begin{minipage}{3.0cm}\vspace{0.08cm}S3D-G \vspace{0.08cm}\end{minipage}& ImageNet & 74.7 & 93.4 \\
\begin{minipage}{3.0cm}\vspace{0.08cm}S3D-G Two-stream \vspace{0.08cm}\end{minipage}& ImageNet & 77.2 & 93.0 \\
\begin{minipage}{3.0cm}\vspace{0.08cm}NL I3D \vspace{0.08cm}\end{minipage}& ImageNet & 77.7 & 93.3 \\ \hline
\begin{minipage}{3.0cm}\vspace{0.08cm}\textbf{SDAS-C} \vspace{0.08cm}\end{minipage}& ImageNet & 78.2 & 93.8 \\
\begin{minipage}{3.0cm}\vspace{0.08cm}\textbf{SDAS-C Two-stream} \vspace{0.08cm}\end{minipage}& ImageNet & \textbf{80.1} & \textbf{94.7} \\ \hline
\end{tabular}
\label{tab:k400-sota}
\vspace{-0.15in}
\end{table}

Table \ref{tab:k400-sota} details the top-1 and top-5 accuracy on Kinetics400 validation set. Overall, randomly initialized SDAS-C outperforms all the networks which are trained on Kinetics400 from scratch. In particular, the top-1 accuracy of SDAS-C can achieve 76.5\%, which makes the absolute improvement over the best competitor R(2+1)D by 4.5\%. Our SDAS-C is even superior to S3D-G pre-trained on ImageNet dataset in terms of top-1 accuracy and obtains comparable top-5 accuracy. In addition, we can also pre-train our SDAS-C architecture on ImageNet as proposed in \citep{qiu2017learning}, which pre-learn the 2D spatial kernel on ImageNet and then finetune the whole networks on video data. As such, the top-1 accuracy can be boost up from 76.5\% to 78.2\%, which is higher than best competitor NL I3D pre-trained on ImageNet. Similar performance trends are observed when extending the networks to two-stream structure.

\section{Conclusion}
We have presented Scheduled Differentiable Architecture Search (SDAS) method which aims to automate the architecture design for visual recognition. Particularly, we study the problem of formulating an architecture or a cell as a directed graph and inducing the optimal computational architecture in a scheduled manner. To verify our claim, we have proposed a scheduled scheme which progressively fixes the optimal operation on each edge and changes the topological connection on each node during training, and integrated such scheme into an efficient architecture search framework based on the continuous relaxation of search space. To encode spatio-temporal information in videos, we further enlarge the search space for video recognition by devising several unique operations. Extensive experiments conducted on CIFAR10, Kinetics10, UCF101 and HMDB51 validate our proposal and analysis. SDAS leads to clear improvements over DAS with a $1.4\sim 2.8\times$ speedup. More remarkably, applying the architecture learnt on CIFAR10/Kinetics10 to ImageNet/Kinetics400 successfully outperforms the advanced hand-crafted structure and demonstrates good transferability on both image and video recognition tasks.


%
%

\bibliographystyle{spbasic}      
\bibliography{egbib}   

\begin{thebibliography}{46}
\providecommand{\natexlab}[1]{#1}
\providecommand{\url}[1]{{#1}}
\providecommand{\urlprefix}{URL }
\expandafter\ifx\csname urlstyle\endcsname\relax
  \providecommand{\doi}[1]{DOI~\discretionary{}{}{}#1}\else
  \providecommand{\doi}{DOI~\discretionary{}{}{}\begingroup
  \urlstyle{rm}\Url}\fi
\providecommand{\eprint}[2][]{\url{#2}}

\bibitem[{Baker et~al.(2018)Baker, Gupta, Raskar, and
  Naik}]{baker2018accelerating}
Baker B, Gupta O, Raskar R, Naik N (2018) Accelerating neural architecture
  search using performance prediction. In: ICLR Workshop

\bibitem[{Brock et~al.(2017)Brock, Lim, Ritchie, and Weston}]{brock2017smash}
Brock A, Lim T, Ritchie JM, Weston N (2017) Smash: one-shot model architecture
  search through hypernetworks. arXiv preprint arXiv:170805344

\bibitem[{Cai et~al.(2018)Cai, Chen, Zhang, Yu, and Wang}]{cai2018efficient}
Cai H, Chen T, Zhang W, Yu Y, Wang J (2018) Efficient architecture search by
  network transformation. In: AAAI

\bibitem[{Carreira and Zisserman(2017)}]{carreira2017quo}
Carreira J, Zisserman A (2017) Quo vadis, action recognition? a new model and
  the kinetics dataset. In: CVPR

\bibitem[{Chollet(2017)}]{Chollet2017CVPR}
Chollet F (2017) Xception: Deep learning with depthwise separable convolutions.
  In: CVPR

\bibitem[{Diba et~al.(2017)Diba, Sharma, and Van~Gool}]{diba2017deep}
Diba A, Sharma V, Van~Gool L (2017) Deep temporal linear encoding networks. In:
  CVPR

\bibitem[{Feichtenhofer et~al.(2016)Feichtenhofer, Pinz, and
  Zisserman}]{feichtenhofer2016convolutional}
Feichtenhofer C, Pinz A, Zisserman A (2016) Convolutional two-stream network
  fusion for video action recognition. In: CVPR

\bibitem[{Ghanem et~al.(2018)Ghanem, Niebles, Snoek, Heilbron, Alwassel,
  Escorcia, Khrisna, Buch, and Dao}]{ghanem2018activitynet}
Ghanem B, Niebles JC, Snoek C, Heilbron FC, Alwassel H, Escorcia V, Khrisna R,
  Buch S, Dao CD (2018) The activitynet large-scale activity recognition
  challenge 2018 summary. arXiv preprint arXiv:180803766

\bibitem[{Hara et~al.(2018)Hara, Kataoka, and Satoh}]{hara2018can}
Hara K, Kataoka H, Satoh Y (2018) Can spatiotemporal 3d cnns retrace the
  history of 2d cnns and imagenet. In: CVPR

\bibitem[{He et~al.(2016)He, Zhang, Ren, and Sun}]{he2015deep}
He K, Zhang X, Ren S, Sun J (2016) Deep residual learning for image
  recognition. In: CVPR

\bibitem[{Howard et~al.(2017)Howard, Zhu, Chen, Kalenichenko, Wang, Weyand,
  Andreetto, and Adam}]{howard2017mobilenets}
Howard AG, Zhu M, Chen B, Kalenichenko D, Wang W, Weyand T, Andreetto M, Adam H
  (2017) Mobilenets: Efficient convolutional neural networks for mobile vision
  applications. arXiv preprint arXiv:170404861

\bibitem[{Hu et~al.(2018)Hu, Shen, and Sun}]{hu2018squeeze}
Hu J, Shen L, Sun G (2018) Squeeze-and-excitation networks. In: CVPR

\bibitem[{Huang et~al.(2017)Huang, Liu, Van Der~Maaten, and
  Weinberger}]{huang2017densely}
Huang G, Liu Z, Van Der~Maaten L, Weinberger KQ (2017) Densely connected
  convolutional networks. In: CVPR

\bibitem[{Ioffe and Szegedy(2015)}]{ioffe2015batch}
Ioffe S, Szegedy C (2015) Batch normalization: Accelerating deep network
  training by reducing internal covariate shift. arXiv preprint arXiv:150203167

\bibitem[{Ji et~al.(2013)Ji, Xu, Yang, and Yu}]{ji20133d}
Ji S, Xu W, Yang M, Yu K (2013) 3d convolutional neural networks for human
  action recognition. IEEE Trans on PAMI 35(1):221--231

\bibitem[{Jia et~al.(2014)Jia, Shelhamer, Donahue, Karayev, Long, Girshick,
  Guadarrama, and Darrell}]{jia2014caffe}
Jia Y, Shelhamer E, Donahue J, Karayev S, Long J, Girshick R, Guadarrama S,
  Darrell T (2014) Caffe: Convolutional architecture for fast feature
  embedding. In: ACM MM

\bibitem[{Karpathy et~al.(2014)Karpathy, Toderici, Shetty, Leung, Sukthankar,
  and Fei-Fei}]{karpathy2014large}
Karpathy A, Toderici G, Shetty S, Leung T, Sukthankar R, Fei-Fei L (2014)
  Large-scale video classification with convolutional neural networks. In: CVPR

\bibitem[{Krizhevsky et~al.(2009)Krizhevsky, Hinton
  et~al.}]{krizhevsky2009learning}
Krizhevsky A, Hinton G, et~al. (2009) Learning multiple layers of features from
  tiny images. Tech. rep., Citeseer

\bibitem[{Krizhevsky et~al.(2012)Krizhevsky, Sutskever, and
  Hinton}]{krizhevsky2012imagenet}
Krizhevsky A, Sutskever I, Hinton GE (2012) Imagenet classification with deep
  convolutional neural networks. In: NIPS

\bibitem[{Kuehne et~al.(2011)Kuehne, Jhuang, Garrote, Poggio, and
  Serre}]{HMDB51}
Kuehne H, Jhuang H, Garrote E, Poggio T, Serre T (2011) {HMDB}: a large video
  database for human motion recognition. In: ICCV

\bibitem[{Liu et~al.(2017)Liu, Zoph, Shlens, Hua, Li, Fei-Fei, Yuille, Huang,
  and Murphy}]{liu2017progressive}
Liu C, Zoph B, Shlens J, Hua W, Li LJ, Fei-Fei L, Yuille A, Huang J, Murphy K
  (2017) Progressive neural architecture search. arXiv preprint arXiv:171200559

\bibitem[{Liu et~al.(2018{\natexlab{a}})Liu, Zoph, Neumann, Shlens, Hua, Li,
  Fei-Fei, Yuille, Huang, and Murphy}]{liu2018progressive}
Liu C, Zoph B, Neumann M, Shlens J, Hua W, Li LJ, Fei-Fei L, Yuille A, Huang J,
  Murphy K (2018{\natexlab{a}}) Progressive neural architecture search. In:
  ECCV

\bibitem[{Liu et~al.(2018{\natexlab{b}})Liu, Simonyan, Vinyals, Fernando, and
  Kavukcuoglu}]{liu2018hierarchical}
Liu H, Simonyan K, Vinyals O, Fernando C, Kavukcuoglu K (2018{\natexlab{b}})
  Hierarchical representations for efficient architecture search. In: ICLR

\bibitem[{Liu et~al.(2019)Liu, Simonyan, and Yang}]{liu2018darts}
Liu H, Simonyan K, Yang Y (2019) Darts: Differentiable architecture search

\bibitem[{Pham et~al.(2018)Pham, Guan, Zoph, Le, and Dean}]{pham2018efficient}
Pham H, Guan MY, Zoph B, Le QV, Dean J (2018) Efficient neural architecture
  search via parameter sharing. In: ICML

\bibitem[{Qiu et~al.(2017)Qiu, Yao, and Mei}]{qiu2017learning}
Qiu Z, Yao T, Mei T (2017) Learning spatio-temporal representation with
  pseudo-3d residual networks. In: ICCV

\bibitem[{Real et~al.(2018)Real, Aggarwal, Huang, and Le}]{real2018regularized}
Real E, Aggarwal A, Huang Y, Le QV (2018) Regularized evolution for image
  classifier architecture search. arXiv preprint arXiv:180201548

\bibitem[{Russakovsky et~al.(2015)Russakovsky, Deng, Su, Krause, Satheesh, Ma,
  Huang, Karpathy, Khosla, Bernstein et~al.}]{russakovsky2015imagenet}
Russakovsky O, Deng J, Su H, Krause J, Satheesh S, Ma S, Huang Z, Karpathy A,
  Khosla A, Bernstein M, et~al. (2015) Imagenet large scale visual recognition
  challenge. IJCV 115(3):211--252

\bibitem[{Simonyan and Zisserman(2014)}]{simonyan2014two}
Simonyan K, Zisserman A (2014) Two-stream convolutional networks for action
  recognition in videos. In: NIPS

\bibitem[{Simonyan and Zisserman(2015)}]{Simonyan:ICLR15}
Simonyan K, Zisserman A (2015) Very deep convolutional networks for large-scale
  image recognition. In: ICLR

\bibitem[{Soomro et~al.(2012)Soomro, Zamir, and Shah}]{UCF101}
Soomro K, Zamir AR, Shah M (2012) {UCF101}: A dataset of 101 human action
  classes from videos in the wild. CRCV-TR-12-01

\bibitem[{Szegedy et~al.(2015)Szegedy, Liu, Jia, Sermanet, Reed, Anguelov,
  Erhan, Vanhoucke, and Rabinovich}]{szegedy2015going}
Szegedy C, Liu W, Jia Y, Sermanet P, Reed S, Anguelov D, Erhan D, Vanhoucke V,
  Rabinovich A (2015) Going deeper with convolutions. In: CVPR

\bibitem[{Tran et~al.(2015)Tran, Bourdev, Fergus, Torresani, and
  Paluri}]{tran2015learning}
Tran D, Bourdev L, Fergus R, Torresani L, Paluri M (2015) Learning
  spatiotemporal features with 3d convolutional networks. In: ICCV

\bibitem[{Tran et~al.(2018)Tran, Wang, Torresani, Ray, LeCun, and
  Paluri}]{tran2018closer}
Tran D, Wang H, Torresani L, Ray J, LeCun Y, Paluri M (2018) A closer look at
  spatiotemporal convolutions for action recognition. In: CVPR

\bibitem[{Wang et~al.(2016)Wang, Xiong, Wang, Qiao, Lin, Tang, and
  Van~Gool}]{wang2016temporal}
Wang L, Xiong Y, Wang Z, Qiao Y, Lin D, Tang X, Van~Gool L (2016) Temporal
  segment networks: Towards good practices for deep action recognition. In:
  ECCV

\bibitem[{Wang et~al.(2018{\natexlab{a}})Wang, Xiong, Wang, Qiao, Lin, Tang,
  and Van~Gool}]{wang2018temporal}
Wang L, Xiong Y, Wang Z, Qiao Y, Lin D, Tang X, Van~Gool L (2018{\natexlab{a}})
  Temporal segment networks for action recognition in videos. IEEE Trans on
  PAMI

\bibitem[{Wang et~al.(2018{\natexlab{b}})Wang, Girshick, Gupta, and
  He}]{wang2018non}
Wang X, Girshick R, Gupta A, He K (2018{\natexlab{b}}) Non-local neural
  networks. In: CVPR

\bibitem[{Xie et~al.(2017)Xie, Girshick, Doll{\'a}r, Tu, and
  He}]{xie2017aggregated}
Xie S, Girshick R, Doll{\'a}r P, Tu Z, He K (2017) Aggregated residual
  transformations for deep neural networks. In: CVPR

\bibitem[{Xie et~al.(2018)Xie, Sun, Huang, Tu, and Murphy}]{xie2018rethinking}
Xie S, Sun C, Huang J, Tu Z, Murphy K (2018) Rethinking spatiotemporal feature
  learning: Speed-accuracy trade-offs in video classification. In: ECCV

\bibitem[{Xie et~al.(2019)Xie, Zheng, Liu, and Lin}]{xie2018snas}
Xie S, Zheng H, Liu C, Lin L (2019) {SNAS}: stochastic neural architecture
  search. In: ICLR

\bibitem[{Yu and Koltun(2016)}]{yu2016multi}
Yu F, Koltun V (2016) Multi-scale context aggregation by dilated convolutions.
  In: ICLR

\bibitem[{Zach et~al.(2007)Zach, Pock, and Bischof}]{zach2007duality}
Zach C, Pock T, Bischof H (2007) A duality based approach for realtime tv-l1
  optical flow. Pattern Recognition

\bibitem[{Zhang et~al.(2018)Zhang, Zhou, Lin, and Sun}]{zhang2018shufflenet}
Zhang X, Zhou X, Lin M, Sun J (2018) Shufflenet: An extremely efficient
  convolutional neural network for mobile devices. In: CVPR

\bibitem[{Zhu et~al.(2016)Zhu, Hu, Sun, Cao, and Qiao}]{zhu2016key}
Zhu W, Hu J, Sun G, Cao X, Qiao Y (2016) A key volume mining deep framework for
  action recognition. In: CVPR

\bibitem[{Zoph and Le(2017)}]{zoph2017neural}
Zoph B, Le QV (2017) Neural architecture search with reinforcement learning.
  In: ICML

\bibitem[{Zoph et~al.(2018)Zoph, Vasudevan, Shlens, and Le}]{zoph2018learning}
Zoph B, Vasudevan V, Shlens J, Le QV (2018) Learning transferable architectures
  for scalable image recognition. In: CVPR

\end{thebibliography}

\end{document}